\documentclass[journal]{IEEEtran}
\usepackage{hyperref}
\usepackage{amsmath}
\usepackage{cite}
\usepackage{algorithm}
\usepackage{algorithmic}
\usepackage{setspace}
\usepackage{graphicx}

\usepackage{multirow}
\usepackage{diagbox}
\usepackage{booktabs}
\usepackage{rotating}
\usepackage{color}
\usepackage{subfig}
\usepackage{amssymb}
\usepackage{tikz,xcolor,hyperref}

\definecolor{lime}{HTML}{A6CE39}
\DeclareRobustCommand{\orcidicon}{
\begin{tikzpicture}
\draw[lime, fill=lime] (0,0)
circle[radius=0.16]
node[white]{{\fontfamily{qag}\selectfont \tiny \.{I}D}}; 
\end{tikzpicture}
\hspace{-2mm}
}
\foreach \x in {A, ..., Z}{%
\expandafter\xdef\csname orcid\x\endcsname{\noexpand\href{https://orcid.org/\csname orcidauthor\x\endcsname}{\noexpand\orcidicon}}
}

\graphicspath{{images/}}
\usepackage[pagewise]{lineno}

\usepackage{hyperref}
\hypersetup{hidelinks,
	colorlinks=true,
	allcolors=blue,
	pdfstartview=Fit,
	breaklinks=true}
\makeatletter
\renewcommand{\maketag@@@}[1]{\hbox{\m@th\normalsize\normalfont#1}}%
\makeatother

\ifCLASSINFOpdf
\else
\fi
\begin{document}
%

		%
		\title{Localized Sparse Incomplete Multi-view\\ Clustering}
		%
		%
		%
		
		\author{Chengliang Liu\hspace{-1.5mm}\orcidA{},
			Zhihao Wu,
			Jie Wen*\hspace{-1.5mm}\orcidB{},
			Yong Xu*,~\IEEEmembership{Senior Member,~IEEE},
			Chao Huang
			\thanks{This work is supported in part by the Shenzhen Science and Technology Program under Grant no. RCBS20210609103709020 and in part by the Shenzhen Science and Technology Innovation Committee under Grant no. GJHZ20210705141812038.}
			\thanks{Chengliang Liu, Zhihao Wu, Jie Wen, and Chao Huang are with the Shenzhen Key Laboratory of Visual Object Detection and Recognition, Harbin Institute of Technology, Shenzhen, Shenzhen 518055, China (Email: liucl1996@163.com; horatio\_ng@163.com; jiewen\_pr@126.com; huangchao\_08@126.com).}
			\thanks{Yong Xu is with the Shenzhen Key Laboratory of Visual Object Detection and Recognition, Harbin Institute of Technology, Shenzhen, Shenzhen 518055, China; is also with the Pengcheng Laboratory, Shenzhen 518055, China. (Email: yongxu@ymail.com).}
			\thanks{Corresponding author: Jie Wen and Yong Xu (Email: jiewen\_pr@126.com; yongxu@ymail.com).}}
		
		%
		%

	\markboth{Journal of \LaTeX\ Class Files,~Vol.~14, No.~8, August~2015}%
	{Shell \MakeLowercase{\textit{et al.}}: Localized Sparse Incomplete Multi-view Clustering}
	%



	\maketitle
	
	\begin{abstract}
		Incomplete multi-view clustering, which aims to solve the clustering problem on the incomplete multi-view data with partial view missing, has received more and more attention in recent years. Although numerous methods have been developed, most of the methods either cannot flexibly handle the incomplete multi-view data with arbitrary missing views or do not consider the negative factor of information imbalance among views. Moreover, some methods do not fully explore the local structure of all incomplete views. To tackle these problems, this paper proposes a simple but effective method, named localized sparse incomplete multi-view clustering (LSIMVC). Different from the existing methods, LSIMVC intends to learn a sparse and structured consensus latent representation from the incomplete multi-view data by optimizing a sparse regularized and novel graph embedded multi-view matrix factorization model. Specifically, in such a novel model based on the matrix factorization, a $l_1$ norm based sparse constraint is introduced to obtain the sparse low-dimensional individual representations and the sparse consensus representation. Moreover, a novel local graph embedding term is introduced to learn the structured consensus representation. Different from the existing works, our local graph embedding term aggregates the graph embedding task and consensus representation learning task into a concise term. Furthermore, to reduce the imbalance factor of incomplete multi-view learning, an adaptive weighted learning scheme is introduced to LSIMVC. Finally, an efficient optimization strategy is given to solve the optimization problem of our proposed model. Comprehensive experimental results performed on six incomplete multi-view databases verify that the performance of our LSIMVC is superior to the state-of-the-art IMC approaches. The code is available in \textit{\href{https://github.com/justsmart/LSIMVC}{https://github.com/justsmart/LSIMVC}}.
	\end{abstract}
	\begin{IEEEkeywords}
		Incomplete multi-view clustering, common latent representation, matrix factorization, graph embedding.
	\end{IEEEkeywords}

	%
	\IEEEpeerreviewmaketitle

	\section{Introduction}
	\label{chap:intro}
	%
	%
	%
	%
	\IEEEPARstart{R}{eal-world} data has increased dramatically not only in scale but also in modalities in recent years. Just like in daily life, we are used to describing a person from different aspects, namely height, weight, grades, income, and so on. Multi-view data from various perspectives or sources is able to represent richer semantic features of observation objects than single-view data \cite{li2021consensus,lin2021completer,li2009music}. In the area of data analysis, multi-view clustering has aroused extensive research enthusiasm \cite{yang2017discrete,chao2021survey,chao2022incomplete,wang2015robust}, and researchers have developed a large amount of multi-view clustering methods in the last few years. For instance, multi-view k-means inspired by single-view k-means exploits the centroid matrices belonging to all views to jointly learn a consistent clustering indicator matrix \cite{cai2013multi}. Originated from single-view spectral clustering method, Li et al. proposed an efficient multi-view spectral clustering method, which focuses on constructing the bipartite graph with respect to the salient points and original points \cite{li2015large}. Another multi-view subspace clustering based method focuses on obtaining the consensus representation from the Laplacian graphs of all views, in which the Laplacian graphs are adaptively constructed from every view via the representation based self-reconstruction technique \cite{gao2015multi}.
	
	Remarkably, existing multi-view clustering methods are based on such an ideal case, where each sample has corresponding feature instances in all views. Whereas, in real practice, it might be difficult to collect the complete multi-view data whose all views are fully observed for all samples. For example, text, picture, video or audio record may be missing to varying degrees when reviewing course content \cite{chao2016consensus,guo2018partial}. In the process of medical data collection, doctors will prescribe different examinations for patients according to different conditions, which means that it is difficult for us to obtain comprehensive examination data, such as magnetic resonance imaging (MRI) data and blood test reports \cite{luo2021MVDRNet}. Incomplete multi-view data brings great challenges to effective clustering \cite{li2014partial,wen2020generalized}. From the perspective of samples, the difference in the number of the available instances for all views directly leads to an imbalance in the total amount of information among samples and views. From the perspective of views, the unpaired missing views dramatically weaken the complementary information among multiple views, especially in the case where massive instances are missing. Therefore, incomplete multi-view clustering (IMC) needs to divide the samples into the corresponding categories as much as possible with the condition of insufficient and unbalanced semantic information, which is a challenging task in comparison with the conventional complete multi-view clustering methods.
	
	Doubtlessly, traditional multi-view clustering algorithms are powerless when facing incomplete multi-view data. In this situation, a few IMC approaches have been developed to cope with the above problems. For instance, the kernel canonical correlation analysis with incomplete views (KCCA-IV) based method \cite{rai2010multiview} calculates the correlation of each view in the projection space, by restoring the kernel matrix corresponding to the incomplete view. Whereas, one obvious drawback is that the KCCA-IV needs at least one intact view with all instances. Many multiple kernel learning based methods \cite{liu2019efficient,liu2019multiple} were proposed, which redefine IMC as a joint mission (\textit{i.e.}, clustering and kernel matrix imputation tasks). However, these methods depend greatly on the quality of the kernels. Another strategy based on matrix factorization is gaining popularity. For example, Li et al. proposed partial multi-view clustering (PMVC) \cite{li2014partial} that is committed to constructing the unified representation of different views in a common latent subspace by combining non-negative matrix factorization (NMF) technique. Incomplete multi-modal visual data grouping (IMG) \cite{zhao2016incomplete} imposes the graph Laplacian constraint on the common representation to reserve the proximity relations within multiple views while getting rid of non-negative constraints. Partial multi-view subspace clustering (PMSC) \cite{xu2018partial}, which is also based on PMVC like IMG, seeks to indirectly learn the self-representation from the public latent subspace instead of the initial data, expecting to reduce the impact of low-quality data on self-representation. However, these methods clearly require that part of the samples contain all view information, and another part contains only one view. To tackle the more general case of missing data, multiple incomplete views clustering (MIC) \cite{shao2015multiple} forces the learning of the latent representation of all instances (including missing instances) on each view and then aggregates all individual representations to consensus latent subspace representations across all views based on the known missing distribution. Similar to MIC, online multi-view clustering (OMVC) \cite{shao2016online} tends to fill the incomplete views with mean vectors, and then dynamically weights each instance in learning a common feature representation according to informative estimations. However, the filling strategy adopted by the above two methods may introduce additional noise. Doubly aligned incomplete multi-view clustering (DAIMC) \cite{hu2018doubly} proposes a dual alignment consensus representation learning model based on the weighted matrix decomposition, that is, in addition to aligning the latent feature spaces of different views, the basis matrices are also aligned based on regularized regression.
	
	In recent years, some view recovery based IMC methods have been designed, which provide a new strategy to address the partial multi-view learning problem. Representatively, Wang et al. provided a cycle generative adversarial network (GAN) \cite{wang2018partial} to recover the missing features and obtain the consensus clustering result simultaneously. Another deep method named COMPLETER \cite{lin2021completer} attempts to directly maximize the two views' information entropies and their mutual information to obtain more informative representation, and a dual prediction mechanism is designed to recover the missing views. Wen et al. proposed a unified embedding alignment framework named UEAF \cite{wen2019unified}, which retains the localized structure across multi-view in subspace learning by introducing a reversal graph learning model. Whereas, the UEAF assumes that the data dimension is much larger than the number of categories; the COMPLETER can only handle two-view data; and the cyclic GAN model requires sufficient paired data to learn cluster centroids. These restrictions on data greatly limit their practical applications.
	
	To remedy aforementioned issues, we creatively propose an IMC method, named localized sparse incomplete multi-view clustering (LSIMVC) in this work. Different from many existing IMC methods, including PMVC, which have some requirements on data distribution, LSIMVC can cope with the incomplete multi-view data with all kinds of view-missing cases, where both the views and the samples can be incomplete. Specifically, the overall objective model of our LSIMVC consists of three major parts: individual sparse representation learning, adaptive multi-view learning, and graph regularized consensus representation learning. Most notably, for the designed graph embedded consensus representation learning model, we cleverly integrate the graph embedding term and consensus representation learning term into one term, which is very different from the existing works. With these three components, LSIMVC is empowered to obtain a sparse and structured consensus representation from the partial multi-view dataset for clustering. Overall, we can summarize our major contributions as follows:
	
	(1) We propose a novel and flexible IMC model named LSIMVC, which is able to cope with all types of partial multi-view data. And sufficient experimental results on six databases illustrate that LSIMVC performs better than other state-of-the-art methods.
	
	(2) In our method, a novel and effective graph regularized consensus representation learning term is designed for IMC. Specifically, very different from the existing works which use two individual terms for consensus representation learning and the exploration of local structure information of data, our graph regularized consensus representation learning term incorporates the above two tasks into one concise term. Moreover, the graph constraint is imposed on the discrepancy between the unified representation and individual representations, which is beneficial to obtain a more discriminative unified representation for a better performance.
	
	We arrange the remaining parts of our paper as follows: In Section \ref{chap:rela}, two typical IMC methods are briefly introduced: KCCA-IV and PMVC, which are related to our work. Section \ref{chap:meth} first describes the proposed LSIMVC, optimization procedure, and finally further analyzes the algorithm complexity. In Section \ref{chap:expe}, comprehensive experiments confirm the effectiveness of the proposed approach. Section \ref{chap:Cond} gives a succinct conclusion to our work.
	\begin{figure*}[t]
		\centering
		\includegraphics[height = 3in, width=7in]{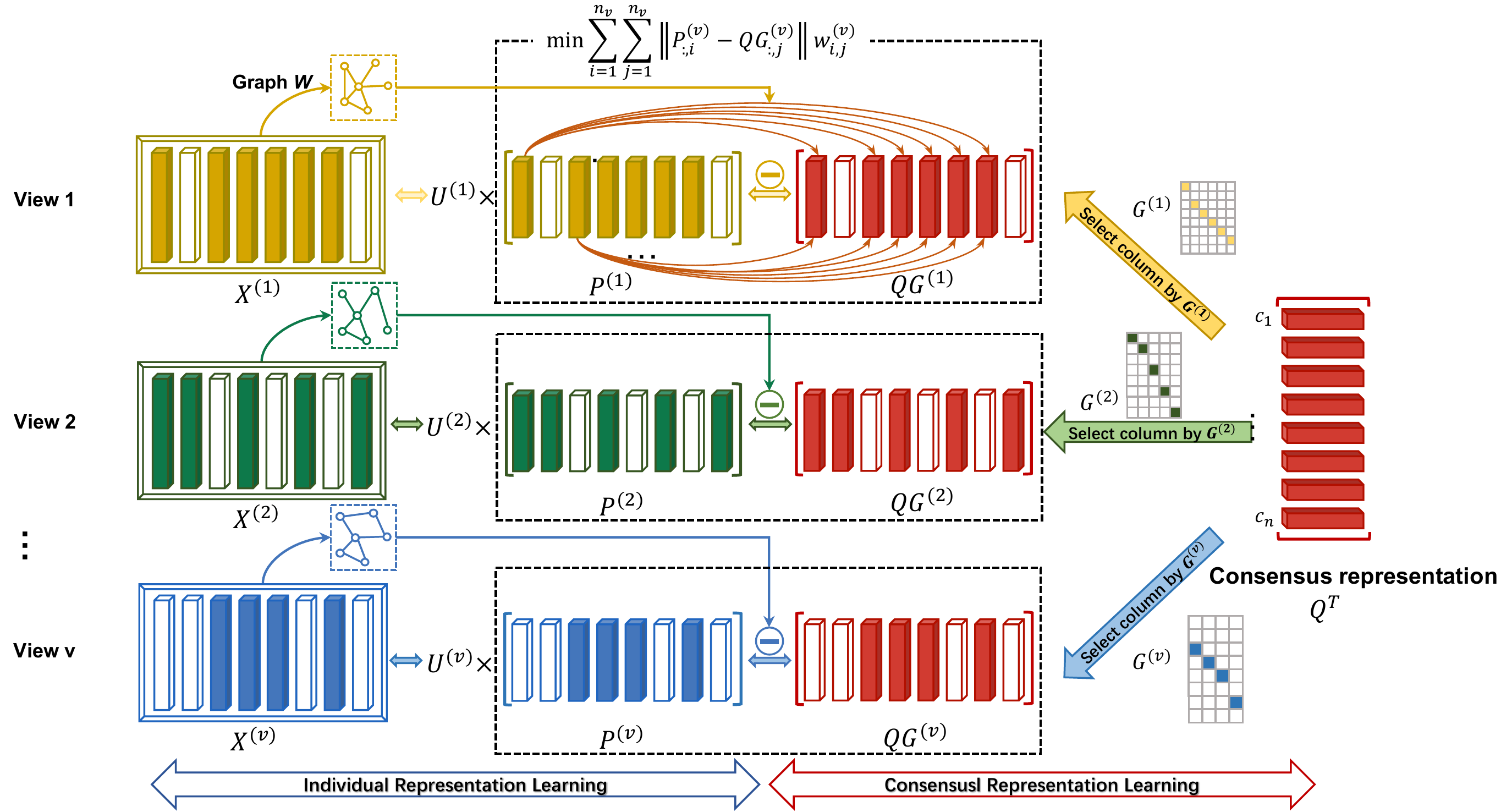}
		\caption{LSIMVC schematic structure. $X^{(v)}$,  $U^{(v)}$ and $P^{(v)}$ represent the original input feature, low-dimensional subspace basis and individual subspace representation of the view $v$, respectively. Q is the common consistency representation, and $G^{(v)}$ is the pre-constructed index matrix based on the $v$th view's prior missing information.}
		\label{fig:f1}
	\end{figure*}
	\section{Related works and preliminary}
	\label{chap:rela}
	Some studies have tried to deal with IMC from multiple perspectives. In this section, we first define some global notations, and then introduce two related methods, namely KCCA-IV \cite{rai2010multiview} and PMVC \cite{li2014partial}. Due to these two methods holding the specific limitations on the incompleteness of samples, we will elaborate further below.
	
	\subsection{Notations}
	For the convenience of introduction, we define some frequently-used symbols in this subsection. We let $X^{(v)}\in \mathbb{R}^{m_{v}\times n_{v}} = [x_{1}^{(v)},x_{2}^{(v)},...,x_{n_{v}}^{(v)}]$ denote the input data of the $v$-th view. $n_{v}$ is the number of available instances in view $v$ and $m_v$ is the dimension of the $v$-th view. $X=\left\{X^{(v)}\right\}_{v=1}^{l}$ denotes the dataset, where $l$ is the number of views. $P^{(v)}$ and $U^{(v)}$ are the low-dimensional representation matrix and basis matrix \textit{w.r.t} view $v$, respectively. And $Q$ is the consensus representation matrix of all views. $G^{(v)}$ represents the prior missing indicator matrix of the $v$-th view and $P^{(v)}_{:,i}$ is the vector corresponding to the $i$-th sample of $P^{(v)}$. Besides, we define operations $\bigg\{\big\|A \big\|_{F} = \sqrt{\sum\limits_{i=1}^{a}\sum\limits_{j=1}^{b}\big|A_{i,j}\big|^2}, A\in \mathbb{R}^{a\times b}\bigg\} $as the Frobenius norm of matrix $A$ and $\big\|A \big\|_{1} = \sum\limits_{i=1}^{a}\sum\limits_{j=1}^{b}|A_{i,j}|$ as the $l_1$ norm of matrix $A$\cite{peng2012rasl,liu2016blessing}.
		
	\subsection{Kernel CCA with incomplete views (KCCA-IV)}
	Canonical correlation analysis (CCA) \cite{hardoon2003kcca,hardoon2004canonical} is a popular method for analyzing correlations between two groups of variables, but it cannot handle nonlinear data. To deal with lots of non-linear data in the real-world, the kernel CCA first utilizes the kernel method to map the features to the high-dimensional space and then maximizes the correlation of different views in the high-dimensional space \cite{shawe2004kernel,blaschko2008semi,gonen2014localized}. For incomplete view data, the KCCA-IV requires at least one complete view (here we assume $X^{(1)}$ is a complete view with $n_c+n_{m}$ available instances and $X^{(2)}$ is the incomplete view with $n_c$ available instances) as basic information for incomplete kernel recovery. In KCCA-IV, the kernel matrix $K_{2}$ corresponding to $X^{(2)}$ can be constructed as follows:
	\begin{equation}
		\label{eq.re1}
		K_{2}=
		\begin{bmatrix}
			K_{2}^{n_cn_c} & K_{2}^{n_cn_m} \\
			\left(K_{2}^{n_cn_m}\right)^{T} & K_{2}^{n_mn_m}
		\end{bmatrix},
	\end{equation}
	where $n_c$ and $n_m$ are the number of available instances and missing instances in $X^{(2)}$, respectively. Next, the KCCA-IV reconstructs the complete $K_{2}$ by Laplacian regularization:
	\begin{equation}
		\label{eq.re2}
		\begin{aligned}
			&\min\limits_{K_{2}\succeq 0}tr(L_{1}K_2)\\
			&s.t.K_{2}(i,j)=K^{n_cn_c}_{2}(i,j), \forall 1 \le (i,j) \le n_c,
		\end{aligned}
	\end{equation}
	where $L_{1} = D_{1} - K_{1}$, and $D_{1}$ is a diagonal matrix consisting of the sum of each row in $K_{1}$. $K^{n_cn_c}_{2} \in \mathbb{R}^{n_c \times n_c}$ (\emph{i.e.}, the upper left sub-block of $K_{2}$) is the kernel matrix corresponding to the available instances of $X^{(2)}$. After obtaining the complete $K_{2}$ (including all $n_c+n_m$ instances), this approach performs the kernel CCA based on the aligned $K_{1}$ and $K_{2}$ to extract consistent low-dimensional features for subsequent clustering. Obviously, the KCCA-IV is powerless in the case where all views are incomplete.
	
	\subsection{Partial multi-view clustering}
	PMVC is another popular method for IMC, which is based on NMF \cite{li2014partial}. This method divides each view into aligned and unaligned parts to construct the consensus representation learning model:
	\begin{equation}
		\small
		\label{eq.re3}
		\begin{aligned}
			&\begin{aligned}
				\min\limits_{P_{c},\hat{P}^{(1)},\hat{P}^{(2)},\atop U^{(1)},U^{(2)}}
				&\left\|
					\left[X_{c}^{(1)},\hat{X}^{(1)}\right]-U^{(1)}\left[P_{c},\hat{P}^{(1)}\right]
				\right\|_{F}^{2}+\lambda\left\|\left[P_{c},\hat{P}^{(1)}\right]
				\right\|_{1}\\
				+&\left\|\left[X_{c}^{(2)},\hat{X}^{(2)}\right]-U^{(2)}\left[P_{c},\hat{P}^{(2)}\right]
				\right\|_{F}^{2}+\lambda\left\|\left[P_{c},\hat{P}^{(2)}\right]
				\right\|_{1}\\
			\end{aligned} \\
			&s.t. U^{(1)} \ge 0, U^{(2)} \ge 0,\hat{P}^{(1)} \ge 0, \hat{P}^{(2)} \ge 0,
		\end{aligned}
	\end{equation}
	where $X^{(1)}=\left[X_{c}^{(1)},\hat{X}^{(1)}\right] \in \mathbb{R}^{m_1\times (n_c+n_{m1})}$ and $X^{(2)}=\left[X_{c}^{(2)},\hat{X}^{(2)}\right] \in \mathbb{R}^{m_2\times (n_c+n_{m2})}$ denote all available instances from two views, respectively. $X^{(1)}_{c} \in \mathbb{R}^{m_1 \times n_{c}}$ and $X^{(2)}_{c} \in \mathbb{R}^{m_2 \times n_{c}}$ denote the paired instances. $\hat{X}^{(1)} \in \mathbb{R}^{m_1\times n_{m1}}$ and $\hat{X}^{(2)} \in \mathbb{R}^{m_2\times n_{m2}}$ denote the unpaired instances of two views. It is no doubt that the total number of samples is $n=n_c+n_{m1}+n_{m2}$. $P_{c} \in \mathbb{R}^{d\times n_c} $ is the common latent representation shared by $n_c$ paired samples across two views, where $d$ represents the dimension of latent common subspace. Similarly, $\hat{P}^{(1)} \in \mathbb{R}^{d\times n_{m1}} $ and $\hat{P}^{(2)} \in \mathbb{R}^{d\times n_{m2}} $ are the low-dimensional representation of the exclusive instance of view 1 and view 2, respectively. The hidden basis matrices of all views are denoted as $U^{(1)} \in \mathbb{R}^{m_1 \times d}$ and $U^{(2)} \in \mathbb{R}^{m_2 \times d}$.

	Since the samples are divided into two parts, PMVC splits the internal structure of each view and learns common low-dimensional representations mainly based on paired samples. Besides, the performance of PMVC is, to a large extent, dependent on the number of available samples on all views. In other words, PMVC is ineffective when there are few or no available shared samples between incomplete views.
	
	\section{The proposed method}
	\label{chap:meth}
	As presented in the related works, PMVC and KCCA-IV are inflexible and have a very strict requirement for the processed incomplete multi-view data. In this section, we provide a novel and flexible IMC model, named LSIMVC, for the arbitrary IMC tasks with all kinds of view-missing cases. The framework of LSIMVC is shown in Fig. \ref{fig:f1}. Different from the other IMC methods, our LSIMVC focuses on learning a sparse and structured consistent latent representation from the partial multi-view data for data clustering.

	\subsection{Learning model of LSIMVC}
	The NMF technique is widely used to obtain the low-dimensional representation of the original data in the field of IMC. According to the statistics, there are two approaches to calculate such a common representation across all views. The one approach directly factorizes the raw multi-view data into a unified representation based on the weighted matrix factorization model with the prior missing-view indicator matrix as weights, where the representative works are OMVC and DAIMC. In another approach, the original multi-view data are decomposed into several individual representations concerning all views and then these individual representations are put towards a consensus representation. The representative works belonging to the second approach are MIC and Graph regularized partial multi-view clustering (GPMVC) \cite{rai2016partial}. Generally speaking, the second approach provides more freedom than the first approach in learning the clustering-friendly consensus representation. Specifically, for the incomplete multi-view data, a naive framework for learning the consensus representation from the individual representations of different views can be expressed as the following matrix factorization model:
	\begin{equation}\label{eq.4}
		\begin{array}{l}
			\mathop {\min }\limits_{\left\{ {{P^{\left( v \right)}},{U^{\left( v \right)}}} \right\}_{v = 1}^l,Q} \sum\limits_{v = 1}^l {\left( \begin{array}{l}
					\left\| {{X^{\left( v \right)}} - {U^{\left( v \right)}}{P^{\left( v \right)}}} \right\|_F^2\\
					+ \lambda \left\| {{P^{\left( v \right)}} - Q{G^{\left( v \right)}}} \right\|_F^2
				\end{array} \right)} \\
			s.t.{\kern 1pt} {\kern 1pt} {U^{\left( v \right)T}}{U^{\left( v \right)}} = I,
		\end{array}
	\end{equation}
	where $X^{(v)} \in \mathbb{R}^{m_v \times n_v}$ represents the existing instances of the $v$-th view, $m_v$ and $n_v$ are the feature dimension and the number of un-missing instances from the \textit{v}-th view, respectively. \textit{l} is the number of views. $U^{(v)} \in \mathbb{R}^{m_v \times c}$ and $P^{(v)} \in \mathbb{R}^{c \times n_v}$ denote the basis matrix for latent subspace and the corresponding low-dimensional representation of the \textit{v}-th view, respectively. $\lambda$ is a positive penalty parameter for the corresponding regularization term. $c$ is the number of clusters and we set $c$ as the feature dimension of the latent subspace in our model. $Q \in \mathbb{R}^{c \times n}$ denotes the low-dimensional consensus representation of samples in common subspace, $n$ is the number of samples including missing and un-missing instances. \textit{I} is the identity matrix. $G^{(v)} \in \mathbb{R}^{n \times n_v}$ is a binary indicator matrix whose elements reflect the instance-missing or instance-available information of the $v$-th view. For the incomplete multi-view data, we can define $G^{(v)} \in \mathbb{R}^{n \times n_v}$ as follows:
	\begin{equation}
		\label{eq.gij}
		G^{(v)}_{i,j}=
		\begin{cases}
			1,  \mbox{\textit{if} $x_{j}^{(v)}$ \textit{is the v-th view of the ith sample}}\\
			0,  \qquad\qquad\qquad\textit{otherwise,}
		\end{cases}
	\end{equation}
	where $x_{j}^{(v)}$ is the \textit{j}-th available instance of $X^{(v)} \in \mathbb{R}^{m_v \times n_v}$ from the \textit{v}-th view.
	
	As shown in (\ref{eq.4}), with the constraint of $G^{(v)}$, the individual representations $P^{(v)}$ of all views can be aggregated into the consensus representation $Q$. And by introducing the orthogonal basis matrix constraint, the trivial solution of $U^{(v)}$ can be avoided. Generally speaking, the practical significance of multi-view clustering comes from the diversity of sample descriptions in the real-world. It is worth noting that different views have different emphasis on the observation of the same object. Moreover, for incomplete multi-view data, the numbers of the available instances in different views may vary greatly. These indicate that the discriminabilities of information from different views may be unbalanced.
	In addition, it is easy to observe that the zero-one label matrix widely used in the pattern classification or clustering tasks is actually a very sparse matrix, which reflects a good distinction between different classes. So it is beneficial to obtain sparse representations for different views. To this end, we transform model (\ref{eq.4}) into the following new model by taking into account the factor of information imbalance among views \cite{nie2016parameter} and sparse representation properties:
	\begin{equation}\label{eq.previous}
		\begin{array}{l}
			\mathop {\min }\limits_{\left\{ {{P^{\left( v \right)}},{U^{\left( v \right)}}} \right\}_{v = 1}^l,Q,\alpha } \sum\limits_{v = 1}^l {\alpha _v^r\left( \begin{array}{l}
					\left\| {{X^{\left( v \right)}} - {U^{\left( v \right)}}{P^{\left( v \right)}}} \right\|_F^2\\
					{\rm{ + }}\beta {\left\| {{P^{\left( v \right)}}} \right\|_1}\\
					+ \lambda \left\| {{P^{\left( v \right)}} - Q{G^{\left( v \right)}}} \right\|_F^2
				\end{array} \right)} \\
			s.t.{\kern 1pt} {\kern 1pt} {U^{\left( v \right)T}}{U^{\left( v \right)}} = I,\sum\limits_{v = 1}^l {{\alpha _v}}  = 1,0 \le {\alpha _v} \le 1,
		\end{array}
	\end{equation}
	where $\alpha_{v}$ is the weight of the \textit{v}-th view. Parameter $r > 1$ is introduced to control the smooth distribution of weights. And $\beta$ is defined as a positive penalty parameter for the corresponding regularization item.

	Importantly, the manifold structure information of individual view can guide the learning of consensus representations effectively, which is widely used in many clustering methods \cite{rai2016partial,yang2017discrete,nie2017multi}. Conventional approach generally imposes a Laplacian constraint on the consensus representation or individual representations of all views based on the nearest neighbor graph. For example, GPMVC introduces the constraint ${\lambda _i}Tr\left( {{P^{\left( v \right)}}{L^{\left( v \right)}}{P^{\left( v \right)T}}} \right)$ to capture the local structures of all views, where $L^{(v)}$ denotes the Laplacian graph based on the prebuilt K-nearest neighbor graph from the present instances of the $v$-th view. Different from this conventional graph embedding constraint, we introduce a new fusion graph embedding constraint on the consensus representation learning term to preserve the local structure of all views. As a result, the new objective function of our LSIMVC is rewritten as follows:
	\begin{equation}
		\label{eq.over}
		\begin{aligned}
			&\min\limits_{\Psi}{\sum\limits_{v=1}^{l}{\alpha_{v}^{r}}}\left(
			\begin{aligned}
				&\left \| X^{(v)}-U^{(v)}P^{(v)}\right \|_{F}^2+\beta\left\|P^{(v)}\right\|_{1}\\
				&+\lambda\sum\limits_{i=1}^{n_v}\sum\limits_{j=1}^{n_v}{\left\|P_{:,i}^{(v)}-QG_{:,j}^{(v)}\right\|_2^2W_{i,j}^{(v)}}
			\end{aligned}
			\right)\\
			&s.t. U^{(v)T}U^{(v)}=I,0 \leq \alpha_{v} \leq 1, \sum_{v=1}^{l}\alpha_{v}=1,
		\end{aligned}
	\end{equation}
	where $\Psi = \left\{\left\{U^{(v)},P^{(v)}\right\}_{v=1}^l,\alpha,Q\right\}$ represents the set of unknown variables. $W^{(v)}$ is a fused graph composed of an identity matrix and a similarity graph whose elements represent the similarity degree of the corresponding two samples. Let $S^{(v)} \in \mathbb{R}^{n_v \times n_v}$ ($S_{i,i}^{(v)}=0$) be the prebuilt similarity graph of the \textit{v}-th view, we can obtain the fused graph $W^{(v)}$ as follows:
	\begin{equation}
		\label{eq.W}
		W^{(v)}=\gamma  S^{(v)}+I,
	\end{equation}
	where $\gamma$ is a fusion weight with positive value.

	Our graph embedding constraint is motivated by the following observation: if instances $x_i^{\left( v \right)}$ and $x_j^{\left( v \right)}$ are nearest neighbors, then their new representations $\big\{P_{:,i}^{\left( v \right)}$, $P_{:,j}^{\left( v \right)}\big\}$ and the corresponding consensus representations $\big\{QG_{:,i}^{\left( v \right)}, QG_{:,j}^{\left( v \right)}\big\}$ should also be nearest neighbors to each other in the subspace. Interestedly, if we set $\gamma = 0$, then model (\ref{eq.over}) will degrade to the previous model (\ref{eq.previous}). With a positive weight $\gamma > 0$, model (\ref{eq.over}) can explore the structure information stored in the graph $S^{(v)}$ to acquire a more discriminative structured representation for clustering. For simplicity, we set $\gamma=1$ in model (\ref{eq.W}). Moreover, it is easy to observe a different point between the proposed graph embedding approach and the conventional graph embedding approach: our graph is imposed on the distance of cross-view-representation (i.e., between each individual representation and the consensus representation) rather than just only on the individual representations or consensus representation of the same sample, which is beneficial to obtain a more discriminative structured consensus representation. In particular, in our work, we simply choose the well-known `Gaussian kernel' to generate the similarity graph $W^{(v)}$, in which the $(i,j)$-th element can be calculated as follows:
	\begin{equation}
	\setlength{\abovedisplayskip}{1pt}
		\label{eq.Wij}
		W_{i,j}^{(v)}=
		\begin{cases}
			1,&i = j \\
			k\left(x_i^{(v)},x_j^{(v)}\right),&i \neq j,
		\end{cases}
	\end{equation}
	where $x_i^{(v)}$ and $x_j^{(v)}$ denote the \textit{i}-th and \textit{j}-th available instance of \textit{v}-th view, $k\!\left(x_i^{(v)},x_j^{(v)}\right)\!\!=\!e^{-\dfrac{\left\|x_i^{(v)}-x_j^{(v)}\right\|^2}{2\sigma^2}}$ is the Gaussian kernel function for measuring the distance between $x_i^{(v)}$ and $x_j^{(v)}$, where $\sigma$ is a scale parameter.

	\subsection{Solution to LSIMVC}
	Problem (\ref{eq.over}) is a representative constrained multi-variables optimization problem concerning variables $ \left\{U^{(v)},P^{(v)}\right\}_{v=1}^l$, $\alpha$, and $Q$. In our work, we provide an iterative algorithm that alternately solves the local optimal solution. And then the difficult optimization problem (\ref{eq.over}) is decomposed into four simple subproblems with respect to these variables as follows:
	
	\textbf{P1} \textit{w.r.t.} $U^{(v)}$: According to the alternating iterative optimization algorithm, we can set variables $\left\{P^{(v)}\right\}_{v=1}^{l}$, $\left\{U^{(k)}\right\}_{k \neq v,k=1}^{l}$, $Q$, and $\alpha$ as the given constants and then solve the following subproblem with respect to variable $U^{(v)}$:
	\begin{equation}
		\label{eq.s1}
		\min\limits_{U^{(v)T}U^{(v)}=I}{\left\|X^{(v)}-U^{(v)}P^{(v)}\right\|_F^2}.
	\end{equation}
	
	According to the orthogonal constraint $U^{(v)T}U^{(v)}=I$, problem (\ref{eq.s1}) can be converted to the following formula:
	\begin{equation}
		\label{eq.s2}
		\max\limits_{U^{(v)T}U^{(v)}=I}{Tr\left(U^{(v)T}X^{(v)}P^{(v)T}\right)}.
	\end{equation}
	
	Following \cite{2006Sparse,2020Generalized}, we can obtain the optimal $U^{(v)}$ as follows:
	\begin{equation}
		\label{eq.s21}
		U^{(v)}=M^{(v)}N^{(v)T}.
	\end{equation}
	where $M^{(v)}$ and $N^{(v)}$ are the singular matrices obtained by conducting the singular value decomposition (SVD) on matrix $X^{(v)}P^{(v)T}$, \emph{i.e.},  $X^{(v)}P^{(v)T} = M^{(v)} \Sigma N^{(v)T}$, where $\Sigma$ is a diagonal matrix whose diagonal elements denote the singular values of matrix $X^{(v)}P^{(v)T}$.
	
	\textbf{P2} \textit{w.r.t.} $P^{(v)}$: Similar to Step \textbf{P1}, we can fix variables $\left\{P^{(k)}\right\}_{k \neq v,k=1}^{l}$, $\left\{U^{(v)}\right\}_{v=1}^{l}$, $Q$, and $\alpha$, and then solve the following subproblem with respect to variable $P^{(v)}$:
	\begin{equation}
		\label{eq.s3}
		\begin{aligned}
			&\min\limits_{P^{(v)}}{\left \| X^{(v)}-U^{(v)}P^{(v)}\right \|_{F}^2+\beta\left\|P^{(v)}\right\|_{1} }\\
			&+\lambda\sum\limits_{i=1}^{n_v}\sum\limits_{j=1}^{n_v}{\left\|P_{:,i}^{(v)}-QG_{:,j}^{(v)}\right\|_2^2W_{i,j}^{(v)}}.
		\end{aligned}
	\end{equation}
	
	For the last term of problem (\ref{eq.s3}), the following equivalent formula satisfies:
	\begin{equation}
		\label{eq.s4}
		\begin{aligned}
			&\sum\limits_{i=1}^{n_v}\sum\limits_{j=1}^{n_v}{\left\|P_{:,i}^{(v)}-QG_{:,j}^{(v)}\right\|_2^2W_{i,j}^{(v)}}\\
			=&\sum\limits_{i=1}^{n_v}\sum\limits_{j=1}^{n_v}\left(
			\begin{aligned}
				&Tr\left(P^{(v)}_{:,i}P^{(v)T}_{:,i}\right)W^{(v)}_{i,j}\\
				&-2Tr\left(P^{(v)}_{:,i}G^{(v)T}_{:,j}Q^{T}\right)W^{(v)}_{i,j}\\
				&+Tr\left(QG^{(v)}_{:,j}G^{(v)T}_{:,j}Q^{T}\right)W^{(v)}_{i,j}
			\end{aligned}
			\right)\\
			=&Tr\left(P^{(v)}D^{(v)}P^{(v)T}\right)+Tr\left(QG^{(v)}D^{(v)}G^{(v)T}Q^{T}\right)\\
			&-2Tr\left(P^{(v)}W^{(v)}G^{(v)T}Q^{T}\right),
		\end{aligned}
	\end{equation}
	where $D^{(v)}$ is a diagonal matrix calculated as $\left(D^{(v)}\right)_{i,i}=\sum\limits_{i=1}^{n_v}W^{(v)}_{i,j}=\sum\limits_{j=1}^{n_v}W^{(v)}_{i,j}$ since graph matrix $W^{(v)}$  has a symmetric structure\cite{zhuang2012non}.
	
	Substituting (\ref{eq.s4}) into (\ref{eq.s3}), the following equivalent problem can be obtained:
	\begin{equation}
		\label{eq.s5}
		\begin{aligned}
			&\min\limits_{P^{(v)}}{\left \| X^{(v)}-U^{(v)}P^{(v)}\right \|_{F}^2+\beta\left\|P^{(v)}\right\|_{1} }\\
			&+\lambda\left(
			\begin{aligned}
				&Tr\left(P^{(v)}D^{(v)}P^{(v)T}\right)\\
				&-2Tr\left(P^{(v)}W^{(v)}G^{(v)T}Q^{T}\right)
			\end{aligned}
			\right)\\
			\Leftrightarrow&\min\limits_{P^{(v)}}Tr\left(X^{(v)}X^{(v)T}\right)+Tr\left(P^{(v)}P^{(v)T}\right)\\
			&-2Tr\left(P^{(v)}X^{(v)T}U^{(v)}\right)+\beta\left\|P^{(v)}\right\|_{1}\\
			&+\lambda\left(
			\begin{aligned}
				&Tr\left(P^{(v)}D^{(v)}P^{(v)T}\right)\\
				&-2Tr\left(P^{(v)}W^{(v)}G^{(v)T}Q^{T}\right)
			\end{aligned}
			\right)\\
			\Leftrightarrow&\min\limits_{P^{(v)}}Tr\left(P^{(v)}\left(I+\lambda D^{(v)}\right)P^{(v)T}\right)+\beta\left\|P^{(v)}\right\|_{1}\\
			&-2Tr\left(P^{(v)}\left(X^{(v)T}U^{(v)}+\lambda W^{(v)}G^{(v)T}Q^{T}\right)\right)\\
			\Leftrightarrow&\min\limits_{P^{(v)}}Tr\left(P^{(v)}H^{(v)}P^{(v)T}\right)+\beta\left\|P^{(v)}\right\|_{1}\\
			&-2Tr\left(P^{(v)}B^{(v)}\right),
		\end{aligned}
	\end{equation}
	where $H^{(v)}=I+\lambda D^{(v)}$ and $B^{(v)}=X^{(v)T}U^{(v)}+\lambda W^{(v)}G^{(v)T}Q^{T}$. Interestingly, we can find that $H^{(v)}$ is also a diagonal matrix with all diagonal elements greater than 0. Therefore, we can equivalently convert problem (\ref{eq.s5}) to problem (\ref{eq.s6}):
	
	\begin{equation}
		\label{eq.s6}
		\begin{aligned}
			&\min\limits_{U^{(v)}}Tr\left(P^{(v)}\left(H^{(v)}\right)^{\frac{1}{2}}P^{(v)T}\right)+\beta\left\|P^{(v)}\right\|_{1}\\
			&-2Tr\left(P^{(v)}\left(H^{(v)}\right)^{\frac{1}{2}}\left(H^{(v)}\right)^{-\frac{1}{2}}B^{(v)}\right)\\
			\Leftrightarrow&\min\limits_{P^{(v)}}{\left\|P^{(v)}\left(H^{(v)}\right)^{\frac{1}{2}}\!-\!B^{(v)T}\left(H^{(v)}\right)^{-\frac{1}{2}}\!\right\|_{F}^{2}}\!\! +\!\beta\left\|P^{(v)}\right\|_{1}\\
			\Leftrightarrow&\min\limits_{P^{(v)}}\sum\limits_{i=1}^{n_v}H^{(v)}_{i,i}\left\|\!\left(\!P^{(v)}_{:,i}\!-\!\left(B^{(v)T}\right)_{:,i}\frac{1}{H^{(v)}_{i,i}}\right)\!\right\|_{2}^{2}\!\!+\!\beta\left\|P^{(v)}_{:,i}\right\|_{1}\\
			\Leftrightarrow&\min\limits_{P^{(v)}}\sum\limits_{i=1}^{n_v}\frac{\beta}{2H^{(v)}_{i,i}}\!\left\|\!P^{(v)}_{:,i}\!\right\|_{1}\!\!+\!\!\frac{1}{2}\!\left\|\!\left(\!P^{(v)}_{:,i}\!-\!\left(B^{(v)T}\right)_{:,i}\frac{1}{H^{(v)}_{i,i}}\!\right)\!\right\|_{2}^{2}\\.
		\end{aligned}
	\end{equation}
	
	Problem (\ref{eq.s6}) can be regarded as $n_v$ sparse regularized optimization problems with respect to all columns of $P^{(v)}$. For each subproblem with respect to any columns of $P^{(v)}$ in problem (\ref{eq.s6}), we can formulate the closed-form solution as follows based on the shrinkage operator\cite{candes2011robust,zhang2018binary}:
	\begin{equation}
		\label{eq.s7}
		\begin{aligned}
			P^{(v)}_{;,i}=&\max\left(0,\frac{\left(B^{(v)T}_{:,i}\right)}{H^{(v)}_{i,i}}-\frac{\beta}{2H^{(v)}_{i,i}}\right)\\
			&+\min\left(0,\frac{\left(B^{(v)T}_{:,i}\right)}{H^{(v)}_{i,i}}+\frac{\beta}{2H^{(v)}_{i,i}}\right),
		\end{aligned}
	\end{equation}
	where $\max{(0,a)}$ sets all elements in vector $a$ less than 0 to 0. $\min{(0,a)}$ sets all elements in vector $a$ larger than 0 to 0.

	\textbf{P3} \textit{w.r.t.} \textit{Q}: Similarly, by ignoring the items that are unrelated to variable $Q$, we can obtain the following subproblem:
	\begin{equation}
		\label{eq.s8}
		\min\limits_{Q}\sum\limits_{v=1}^{l}\alpha^{r}_v\left(\sum\limits_{i=1}^{n_v}\sum\limits_{j=1}^{n_v}\left\|P_{:,i}^{(v)}-QG_{:,j}^{(v)}\right\|_2^2W_{i,j}^{(v)}\right).
	\end{equation}
	
	Substituting (\ref{eq.s4}) into (\ref{eq.s8}) and further removing the irrelevant items to variable $Q$, we obtain:
	\begin{equation}
		\label{eq.s9}
		\begin{aligned}
			&\min\limits_{Q}\sum\limits_{v=1}^{l}\alpha_{v}^{r}\left(
			\begin{aligned}
				&Tr\left(QG^{(v)}D^{(v)}G^{(v)T}Q^{T}\right)\\
				&-2Tr\left(P^{(v)}W^{(v)}G^{(v)T}Q^{T}\right)
			\end{aligned}
			\right)\\
			\Leftrightarrow&\min\limits_{Q}Tr\left(Q\left(\sum\limits_{v=1}^{l}\alpha_{v}^{r}G^{(v)}D^{(v)}G^{(v)T}\right)Q^{T}\right)\\
			&-2Tr\left(Q^{T}\sum\limits_{v=1}^{l}\alpha_{v}^{r}P^{(v)}W^{(v)}G^{(v)T}\right).
		\end{aligned}
	\end{equation}
	
	From problem (\ref{eq.s9}), we can first calculate its partial derivative \textit{w.r.t.} variable $Q$ and then set that partial derivative to zero. As a result, the optimal solution of (\ref{eq.s9}) is:
	\begin{equation}
		\label{eq.s10}
		Q=\sum\limits_{v=1}^{l}\alpha_{v}^{r}P^{(v)}W^{(v)}G^{(v)T}\left(\sum\limits_{v=1}^{l}\alpha_{v}^{r}G^{(v)}D^{(v)}G^{(v)T}\right)^{-1}
	\end{equation}

	\textbf{P4} \textit{w.r.t.} $\alpha$: We first suppose:
	\begin{equation}
		\begin{aligned}
			e_{v}=&\left\|X^{(v)}-U^{(v)}P^{(v)}\right\|_{F}^{2}+\beta\left\|P^{(v)}\right\|_{1}\\
			&+\lambda\sum\limits_{i=1}^{n_v}\sum\limits_{j=1}^{n_v}{\left\|P_{:,i}^{(v)}-QG_{:,j}^{(v)}\right\|_2^2W_{i,j}^{(v)}},
		\end{aligned}
	\end{equation}
	and then we can obtain the following optimization subproblem by fixing all the other variables:
	\begin{equation}
		\label{eq.s11}
		\begin{aligned}
			&\min\limits_{\alpha}\sum\limits_{v=1}^{l}\alpha^{r}_{v}e_{v}\\
			&s.t. 1\ge\alpha_{v}\ge0, \sum\limits_{v=1}^{l}\alpha_{v}=1.
		\end{aligned}
	\end{equation}
	
	For problem (\ref{eq.s11}), it is effortless for us to acquire the following closed-form solution \cite{zhang2018binary}:
	\begin{equation}
		\label{eq.s12}
		\alpha_{v}=\frac{e_{v}^{1/(1-r)}}{\sum\limits_{v=1}^{l}e_{v}^{1/(1-r)}}.
	\end{equation}
	
	According to the popular alternating optimization scheme, by iteratively calculating the above variables via the above four steps, \textit{i.e.}, \textbf{P1}-\textbf{P4}, we can obtain the local optimal solutions for problem (\ref{eq.over}). The whole procedures for optimizing objective problem (\ref{eq.over}) are summarized in the following Algorithm \ref{al.1}.
	\begin{algorithm}
		\caption{LSIMVC(solving (\ref{eq.over}))}
		\label{al.1}
		\textbf{Input}: Incomplete multi-view data $\left\{X^{(v)}\right\}_{v=1}^{l}$ with view-presence information, parameters $\lambda$, $\beta$, and $r$.\\
		\textbf{Initialization}: Construct $\left\{G^{(v)}\right\}_{v=1}^{l}$ according to the view-presence information and construct the similarity graph $\left\{W^{(v)}\right\}_{v=1}^{l}$ according to the k-nearest neighbor graph construction scheme. Initialize $\left\{U^{(v)}\right\}_{v=1}^{l}$ as an orthogonal matrix with random values and set $P^{(v)}=U^{(v)T}X^{(v)}$. Set $\alpha=\emph{1}$ with all elements as 1.
		\begin{algorithmic}
			\WHILE{not converged}
			\STATE 1.Update $Q$ using (\ref{eq.s10}).
			\STATE 2.Update $\left\{U^{(v)}\right\}_{v=1}^{l}$ by solving (\ref{eq.s2}).
			\STATE 3.Update $\left\{P^{(v)}\right\}_{v=1}^{l}$ using (\ref{eq.s7}).
			\STATE 4.Update $\alpha_v$ using (\ref{eq.s12}).
			\ENDWHILE
		\end{algorithmic}
		\textbf{Output}: $P$
	\end{algorithm}
	\subsection{Computational complexity analysis}
	For the optimization algorithm listed in Algorithm \ref{al.1}, its computational complexity is the summarization of the computational complexities with respect to steps \textbf{P1}-\textbf{P4}. For step \textbf{P1}, the major computational operation is SVD on matrix $X^{(v)}P^{(v)T}$, which costs about $O\left(m_{v}c^{2}\right)$ for the \textit{v}-th view. Therefore, the computational complexity of \textbf{P1} for all views is about $O\left(\sum\limits_{v=1}^{l}m_{v}c^{2}\right)$. For \textbf{P2} and \textbf{P4}, it is obvious that there are no  high computational operations in the two steps and thus we can ignore their computational complexities. For \textbf{P3} \textit{w.r.t.} $Q$, there is a matrix inverse operation with the computational complexity of about $O\left(n^{3}\right)$ for the matrix $\sum\limits_{v=1}^{l}\alpha_{v}^{r}G^{(v)}D^{(v)}G^{(v)T}$ with the size of $n \times n$. But interestingly, matrix $\sum\limits_{v=1}^{l}\alpha_{v}^{r}G^{(v)}D^{(v)}G^{(v)T}$ is a positive diagonal matrix indeed and its inverse matrix can be efficiently obtained by computing the reciprocal of the diagonal elements, so we also consider the computational complexity of \textbf{P3} to be negligible. Finally,The total computational complexity of Algorithm \ref{al.1} is about $O\left(t\sum\limits_{v=1}^{l}m_{v}c^{2}\right)$, where $t$  denotes the convergent step or the maximum iteration number.
	
	\section{EXPERIMENTS AND ANALYSIS}
	\label{chap:expe}
	In this section, we compare multiple top-performing IMC methods on diverse datasets to verify the effectiveness of the LSIMVC.
	
	\subsection{Experiment settings}
	\textbf{Compared methods}: We select eleven popular methods in the field of IMC to compare with our model. In addition to MIC \cite{shao2015multiple}, OMVC \cite{shao2016online}, DAIMC \cite{hu2018doubly}, UEAF \cite{wen2019unified}, and COMPLETER \cite{lin2021completer} mentioned in the introduction, the other five methods are as follows:
	
	(1) Best single view (\textbf{BSV}) \cite{zhao2016incomplete}: As a common baseline method, BSV does nothing to the missing instances except populate them with the average instance calculated by other available instances of the same view. Then, \textit{k}-means is performed on each view separately and the optimal clustering result is selected.
	
	(2) \textbf{Concat} \cite{zhao2016incomplete}: Concat is another typical baseline approach based on a naive information fusion strategy, which first aligns all views with the same filling method as the BSV and splices all instances belonging to the same sample directly to get a unified representation.
	
	
	(3) One-pass incomplete multi-view clustering (\textbf{OPIMC}) \cite{hu2019one}: OPIMC attempts to combine regularized matrix factorization and weighted matrix factorization to learn a common clustering indicator matrix (consensus representation), and proposes a chunk optimization method to reduce computational and storage complexity in larger datasets.
	
	(4) Multiple kernel k-means with incomplete kernels (\textbf{MKKM-IK-MKC}) \cite{liu2019multiple}: MKKM-IK-MKC can simultaneously perform kernel completion and clustering instead of two-stage, and it does not need at least one complete kernel.
	
	(5) Perturbation-oriented incomplete multi-view clustering (\textbf{PIC}) \cite{wang2019spectral}: PIC learns the consensus Laplacian matrix of multiple views according to spectral perturbation theory and partitions data by spectral clustering method to get the final clustering result.
	
	(6) Robust multi-view clustering with incomplete information (\textbf{SURE}) \cite{yang2022robust}: SURE is a unified deep framework, based on noise-robust contrastive learning, for view-misalignment and view-incompleteness problems in multi-view clustering. 

	The above comparison methods, except BSV, Concat, COMPLETER, and SURE, belong to three types of typical IMC methods, \textit{i.e.}, multi-kernel learning based method (MKKM-IK-MKC), spectral clustering based method (PIC), and matrix decomposition based method (MIC, OMVC, OPIMC, etc). It should be pointed out that the COMPLETER and SURE are only applicable to the two-view incomplete dataset, so we just evaluate their performance on the Animal dataset. The specific evaluation results are given in the next subsection.
	
	\textbf{Datasets}: In this paper, we select six multi-view datasets to verify the effectiveness of our proposed method. The detailed information of these datasets is listed in Table \ref{tab:1}.
	
	(1) \textbf{\textit{Handwritten digit}} \cite{asuncion2007uci}: The Handwritten digit dataset has gradually become one of the most popular datasets for IMC tasks. It contains 2000 images, with 10 categories from '0' to '9' and each image is observed from six perspectives. In this paper, we select five views, whose dimensions are 76, 216, 64, 240, and 47, to evaluate those approaches.
	
	(2) \textbf{\textit{BBCSport}} \cite{greene2006practical}: The BBCSport dataset comes from the BBCSport website and includes 773 documents covering five sports. And each document is divided into four parts, namely four views. In our experiments, we select 116 documents from the original dataset as our evaluation data.
	
	(3) \textbf{\textit{3 Sources}} \cite{greene2006practical}: The original 3 Sources dataset contains 948 texts divided into six categories, which is from 3 text databases such as the BBC. In the paper, we experiment on a subset with 169 samples covering all three views.
	
	(4) \textbf{\textit{Caltech7}} \cite{cai2013multi}: Caltech7 is derived from Caltech101\cite{fei2004learning}, which contains 1474 images with seven categories, such as 'dollar bill', 'Garfield, 'stop sign', 'Snoopy', 'faces', 'motorbikes', and 'Windsor chair'. All images were featured in six ways, \textit{i.e.}, LBP, Gist \cite{oliva2001modeling}, Hog\cite{felzenszwalb2008discriminatively}, Cenhist, Gabor, and wavelet-moments.
	
	(5) \textbf{\textit{NH\_face}} (Notting Hill face) \cite{cao2015constrained}: As a subset of the NH database derived from the movie 'Notting Hill', NH\_face is composed of 4660 samples from five persons \cite{wu2013constrained}, and each image, with the size of 40$\times$50, is described by three kinds of features, \textit{i.e.}, Gabor, gray pixels features, and LBP.
	
	(6) \textbf{\textit{Animal}} \cite{fei2004learning,zhang2019cpm}: The Animal is a large database with 10158 images across 50 classes. There are two types of features extracted by deep neural networks, \textit{i.e.}, DECAF \cite{krizhevsky2012imagenet} and VGG19 \cite{simonyan2014very}, in the Animal database.
	
	\begin{table}[htbp]
		\centering
		\caption{DETAILED STATISTICS ABOUT SIX MULTI-VIEW DATABASES.}
		\label{tab:1}
		\resizebox{0.49\textwidth}{!}{
			\begin{tabular}{ccccc}
				\toprule
				Database   & \# Class & \# View & \# Samples &      \# Features       \\ \midrule\midrule
				Handwritten &    10    &    5    &    2000    &    76/216/64/240/47    \\
				BBCSport   &    5     &    4    &    116     &  1991/2063/2113/2158   \\
				3 Sources  &    6     &    3    &    169     &     3560/3631/3068     \\
				Caltech7   &    7     &    6    &    1474    & 48/40/254/1984/512/928 \\
				NH\_face   &    5     &    3    &    4660    &     6750/2000/3304     \\
				Animal    &    50    &    2    &   10158    &       4096/4096        \\ \bottomrule
		\end{tabular}}
	\end{table}
	
	\textbf{Incomplete multi-view data construction}: The above six databases are all complete multi-view data. In the experiment, we adopt two approaches to construct incomplete data for evaluation. We randomly select $30\%$, $50\%$, and $70\%$ instances of each view for the Handwritten database and $10\%$, $30\%$, and $50\%$ of each view's instances for the BBCSport, 3 Sources, Caltech7, and NH\_face databases as the missing instances. And we guarantee that at least one view is available for any samples. Since the Animal database contains only two views, it is difficult to build a large proportion of incomplete data (\textit{i.e.}, the maximum missing proportion is $50\%$) while maintaining the same amount of available samples\cite{liu2020efficient}. A better option is to construct a special incompleteness case where some samples are view-complete. We randomly select $30\%$, $50\%$, and $70\%$ samples with the complete view (named paired samples) and randomly delete one view of the remaining samples on the premise that the number of available instances from each view is approximately equal.
	
	\textbf{Evalution}:
	Same as other IMC works \cite{cai2013multi,hu2018doubly,rai2016partial}, we also adopt clustering accuracy (ACC), normalized mutual information (NMI), and purity as metrics to evaluate the performance of the aforementioned methods. Simply put, the higher the values of the metrics, the better the clustering effect. Taking into account the parameter sensitivity of these methods, we iteratively search for the optimal parameters one by one in a given set of parameters. Meanwhile, in order to avoid the randomness of the results to the greatest extent, we randomly generate a group of masks for each original database at each missing rate and calculate the average performance of the methods on these incomplete databases as the final results.
	
	\subsection{Experimental results and analysis}

	\begin{table*}[htbp]
		\small
		\centering
		\caption{METRICS OF TEN IMC MODELS ON THE \textbf{BBCSPORT} AND \textbf{3SOURCES} DATABASES WITH DIFFERENT INCOMPLETE RATIOS. The 1st/2nd BEST RESULTS ARE MARKED IN \textcolor{red}{RED}/\textcolor{blue}{BLUE}.}
		\resizebox{0.95\textwidth}{!}{
			\begin{tabular}{|c|l|ccc|ccc|ccc|}
				\hline
				\multicolumn{2}{|c|}{} & \multicolumn{3}{c|}{ACC (\%)}      & \multicolumn{3}{c|}{NMI (\%)}      & \multicolumn{3}{c|}{Purity (\%)} \\
				\hline
				\multicolumn{1}{|c|}{Dataset} & Method\textbackslash{}Rate & 10\% & 30\% & 50\% &10\% & 30\% & 50\% &10\% & 30\% & 50\% \\
				\hline	
				\multirow{11}{*}{\begin{turn}{0}BBCSport\end{turn}}
				&BSV	&58.62$\pm$3.94	&51.31$\pm$5.33	&44.03$\pm$3.78	&43.73$\pm$7.43	&31.03$\pm$2.08	&21.40$\pm$2.61	&65.79$\pm$5.52	&55.07$\pm$1.51	&47.59$\pm$2.28\\
				&Concat	&70.62$\pm$3.76	&58.72$\pm$5.42	&33.21$\pm$2.19	&61.69$\pm$6.72	&38.92$\pm$7.87	&18.61$\pm$1.44	&80.59$\pm$4.59	&63.24$\pm$5.82	&37.00$\pm$1.54\\

				&MIC	&51.21$\pm$4.21	&46.21$\pm$4.71	&46.03$\pm$5.19	&29.90$\pm$6.25	&25.84$\pm$3.24	&24.01$\pm$5.39	&55.00$\pm$4.15	&51.72$\pm$4.27	&52.41$\pm$6.23\\
				&DAIMC	&68.62$\pm$4.59	&63.45$\pm$10.97	&56.89$\pm$5.59	&56.62$\pm$4.60	&50.17$\pm$9.91	&37.89$\pm$6.22	&76.90$\pm$5.89	&71.72$\pm$10.76	&61.03$\pm$5.08\\
				&OMVC	&53.33$\pm$3.21	&51.38$\pm$3.06	&48.79$\pm$3.10	&30.64$\pm$2.00	&41.57$\pm$2.79	&40.63$\pm$2.45	&56.49$\pm$2.81	&59.20$\pm$2.12	&57.47$\pm$2.80\\
				&OPIMC	&54.14$\pm$4.78	&52.93$\pm$4.93	&45.69$\pm$6.00	&35.66$\pm$4.71	&31.56$\pm$6.10	&21.75$\pm$6.44	&58.28$\pm$4.82	&56.72$\pm$5.76	&50.86$\pm$6.87\\
				&MKKM-IK-MKC	&77.55$\pm$2.01	&75.66$\pm$3.01	&67.07$\pm$3.51	&\textcolor{blue}{72.91$\pm$3.29}	&64.42$\pm$4.69	&53.52$\pm$4.74	&\textcolor{blue}{88.76$\pm$2.01}	&84.03$\pm$3.22	&77.00$\pm$3.55\\
				&PIC	&75.52$\pm$1.57	&74.48$\pm$3.32	&\textcolor{blue}{69.48$\pm$6.02}	&70.94$\pm$2.22	&64.18$\pm$2.74	&53.91$\pm$6.22	&87.41$\pm$1.44	&82.41$\pm$2.48	&\textcolor{blue}{77.14$\pm$6.04}\\
				&UEAF  &\textcolor{blue}{78.22$\pm$0.94} &\textcolor{blue}{77.24$\pm$3.08}  &69.31$\pm$3.43  &70.71$\pm$2.59 &\textcolor{red}{\textbf{68.25$\pm$5.13}} &\textcolor{blue}{55.13$\pm$7.09} &87.41$\pm$2.24 &\textcolor{blue}{87.07$\pm$2.86} &77.07$\pm$3.98\\
				&Ours	&\textcolor{red}{\textbf{80.69$\pm$3.69}}	&\textcolor{red}{\textbf{78.79$\pm$1.68}}	&\textcolor{red}{\textbf{72.07$\pm$2.25}}	&\textcolor{red}{\textbf{73.57$\pm$1.66}}	&\textcolor{blue}{67.71$\pm$4.02}	&\textcolor{red}{\textbf{58.65$\pm$2.33}}	&\textcolor{red}{\textbf{89.83$\pm$1.12}}	&\textcolor{red}{\textbf{87.76$\pm$0.73}}	&\textcolor{red}{\textbf{80.17$\pm$2.20}}\\
				\hline
				\multirow{11}{*}{\begin{turn}{0}3Sources\end{turn}}
				&BSV	&56.90$\pm$3.69	&47.38$\pm$3.07	&39.24$\pm$3.08	&50.07$\pm$1.22	&34.46$\pm$4.07	&22.34$\pm$1.91	&68.14$\pm$1.67	&57.63$\pm$1.32	&48.99$\pm$0.63\\
				&Concat	&53.54$\pm$3.00	&46.79$\pm$3.99	&37.68$\pm$2.91	&51.98$\pm$1.37	&37.87$\pm$3.66	&18.32$\pm$3.25	&69.78$\pm$1.09	&58.51$\pm$3.18	&46.48$\pm$2.82\\

				&MIC	&49.11$\pm$3.60	&47.69$\pm$7.61	&42.49$\pm$8.63	&37.23$\pm$6.13	&38.62$\pm$3.81	&26.08$\pm$7.42	&57.28$\pm$3.36	&61.30$\pm$4.28	&52.31$\pm$4.96\\
				&DAIMC	&56.33$\pm$4.23	&52.43$\pm$6.63	&50.73$\pm$3.87	&52.98$\pm$3.65	&49.07$\pm$5.78	&41.64$\pm$2.43	&68.99$\pm$4.26	&67.21$\pm$4.89	&63.56$\pm$3.38\\
				&OMVC	&43.95$\pm$7.35	&41.11$\pm$4.31	&39.53$\pm$3.63	&36.48$\pm$10.77	&28.42$\pm$3.41	&24.34$\pm$1.50	&59.37$\pm$8.26	&48.76$\pm$5.44	&45.44$\pm$3.10\\
				&OPIMC	&55.73$\pm$2.85	&54.20$\pm$4.48	&43.08$\pm$6.98	&40.62$\pm$2.28	&38.83$\pm$3.86	&22.69$\pm$3.83	&64.73$\pm$1.70	&64.26$\pm$2.03	&53.61$\pm$4.36\\
				&MKKM-IK-MKC	&54.44$\pm$0.78	&49.44$\pm$5.98	&50.39$\pm$4.14	&55.46$\pm$2.55	&48.76$\pm$3.48	&48.17$\pm$2.76	&76.97$\pm$1.06	&70.39$\pm$4.38	&69.59$\pm$2.17   \\
				&PIC	&\textcolor{blue}{71.83$\pm$4.59}	&\textcolor{blue}{70.29$\pm$4.25}	&\textcolor{blue}{55.50$\pm$4.15}	&\textcolor{blue}{67.02$\pm$5.77}	&\textcolor{blue}{63.66$\pm$3.20}	&\textcolor{blue}{52.01$\pm$3.81}	&\textcolor{blue}{80.24$\pm$3.39}	&\textcolor{blue}{79.05$\pm$1.59}	&\textcolor{blue}{73.85$\pm$2.49}   \\
				&UEAF  &62.60$\pm$2.73 & 55.62$\pm$5.69  & 52.78$\pm$4.53  &56.47$\pm$4.37 &52.06$\pm$2.33 &45.19$\pm$6.54 &75.50$\pm$3.20 &71.95$\pm$4.16 &67.69$\pm$5.28\\
				&Ours	&\textcolor{red}{\textbf{74.79$\pm$4.84}}	&\textcolor{red}{\textbf{74.32$\pm$4.54}}	&\textcolor{red}{\textbf{67.57$\pm$4.17}}	&\textcolor{red}{\textbf{67.87$\pm$3.44}}	&\textcolor{red}{\textbf{66.16$\pm$4.05}}	&\textcolor{red}{\textbf{59.01$\pm$3.67}}	&\textcolor{red}{\textbf{83.31$\pm$2.76}}	&\textcolor{red}{\textbf{80.83$\pm$3.66}}	&\textcolor{red}{\textbf{76.33$\pm$3.05}}\\
				\hline
			\end{tabular}
		}
		\label{tab:res1}
	\end{table*}
	
	\begin{table*}[htbp]
		\small
		\centering
		\caption{METRICS OF TEN IMC MODELS ON THE \textbf{CALTECH7} AND \textbf{NH\_FACE} DATABASES WITH DIFFERENT INCOMPLETE RATIOS. The 1st/2nd BEST RESULTS ARE MARKED IN \textcolor{red}{RED}/\textcolor{blue}{BLUE}.}
		
		\resizebox{0.95\textwidth}{!}{
			\begin{tabular}{|c|l|ccc|ccc|ccc|}
				\hline
				\multicolumn{2}{|c|}{} & \multicolumn{3}{c|}{ACC (\%)}      & \multicolumn{3}{c|}{NMI (\%)}      & \multicolumn{3}{c|}{Purity (\%)} \\
				\hline
				\multicolumn{1}{|c|}{Dataset} & Method\textbackslash{}Rate & 10\% & 30\% & 50\% &10\% & 30\% & 50\% &10\% & 30\% & 50\% \\
				\hline
				\multirow{9}{*}{\begin{turn}{0}Caltech7\end{turn}}
				&BSV	&43.89$\pm$1.37	&39.06$\pm$1.26	&38.31$\pm$1.68	&39.66$\pm$2.23	&31.63$\pm$1.51	&26.81$\pm$1.38	&84.08$\pm$1.23	&75.25$\pm$0.71	&68.97$\pm$0.49\\
				&Concat	&41.25$\pm$1.67	&40.55$\pm$1.89	&38.06$\pm$0.88	&43.48$\pm$0.92	&37.99$\pm$2.17	&30.28$\pm$0.66	&\textcolor{blue}{84.91$\pm$0.50}	&82.54$\pm$1.12	&77.56$\pm$0.98\\
				&MIC	&44.07$\pm$4.97	&38.01$\pm$2.12	&35.80$\pm$2.34	&33.71$\pm$2.66	&27.35$\pm$1.69	&20.44$\pm$0.98	&78.12$\pm$1.76	&73.31$\pm$0.72	&68.26$\pm$1.40\\
				&DAIMC	&48.29$\pm$6.76	&47.46$\pm$3.42	&44.89$\pm$4.88	&\textcolor{blue}{44.61$\pm$3.88}	&38.45$\pm$2.88	&36.28$\pm$2.34	&83.32$\pm$1.31	&76.83$\pm$3.23	&75.50$\pm$1.17\\
				&OMVC	&40.88$\pm$1.54	&36.82$\pm$1.65	&33.28$\pm$4.40	&28.13$\pm$2.54	&25.32$\pm$1.03	&18.76$\pm$4.22	&79.21$\pm$1.77	&77.73$\pm$1.35	&74.05$\pm$4.74\\
				&OPIMC	&49.24$\pm$2.89	&48.34$\pm$4.36	&44.12$\pm$5.85	&42.98$\pm$1.02	&41.54$\pm$2.38 &35.98$\pm$2.77	&84.89$\pm$0.69	&83.70$\pm$1.80	&80.64$\pm$2.06\\
				&MKKM-IK-MKC	&36.54$\pm$0.51	&34.87$\pm$1.53	&36.05$\pm$0.45	&24.09$\pm$0.98	&23.45$\pm$0.52	&22.91$\pm$0.67	&72.98$\pm$0.80	&73.82$\pm$0.53	&72.52$\pm$1.55\\
				&PIC	&\textcolor{blue}{58.82$\pm$2.95}	&\textcolor{red}{\textbf{58.24$\pm$1.20}}	&\textcolor{blue}{56.50$\pm$2.93}	&41.73$\pm$3.93	&\textcolor{blue}{44.44$\pm$3.12}	&\textcolor{blue}{43.51$\pm$1.50}	&83.99$\pm$0.54	&\textcolor{blue}{83.89$\pm$0.53}	&\textcolor{blue}{83.64$\pm$0.55}\\
				&UEAF	&50.82$\pm$4.05	&42.71$\pm$0.84	&36.32$\pm$4.22	&39.44$\pm$2.07	&31.07$\pm$1.99	&24.02$\pm$1.37	&81.49$\pm$1.78	&78.26$\pm$2.12	&76.29$\pm$1.93\\

				&Ours	&\textcolor{red}{\textbf{59.62$\pm$4.28}}	&\textcolor{blue}{57.64$\pm$1.68}	&\textcolor{red}{\textbf{56.66$\pm$2.66}}	&\textcolor{red}{\textbf{49.72$\pm$4.27}}	&\textcolor{red}{\textbf{46.19$\pm$1.86}} &\textcolor{red}{\textbf{44.61$\pm$2.17}}	&\textcolor{red}{\textbf{85.10$\pm$0.61}}	&\textcolor{red}{\textbf{84.31$\pm$0.57}}	&\textcolor{red}{\textbf{83.91$\pm$0.57}}   \\
				
				\hline
				\multirow{9}{*}{\begin{turn}{0}NH\_face\end{turn}}
				&BSV	&69.09$\pm$4.76	&56.82$\pm$2.28	&46.54$\pm$1.90	&56.26$\pm$4.07	&39.29$\pm$2.63	&26.20$\pm$1.09	&73.59$\pm$2.96	&60.13$\pm$1.52	&50.15$\pm$1.28\\
				&Concat	&85.87$\pm$2.64	&63.14$\pm$2.78	&52.99$\pm$1.84	&\textcolor{blue}{81.46$\pm$1.70}	&59.12$\pm$1.14	&47.42$\pm$1.29	&\textcolor{blue}{87.39$\pm$1.57}	&\textcolor{blue}{87.39$\pm$1.57}	&62.21$\pm$1.04\\
				&MIC	&78.83$\pm$4.07	&77.22$\pm$0.76	&75.77$\pm$4.05	&73.04$\pm$2.78	&66.82$\pm$0.80	&62.84$\pm$3.20	&82.54$\pm$1.66
				&78.83$\pm$0.64	&77.40$\pm$3.48\\
				&DAIMC	&\textcolor{blue}{87.42$\pm$4.15} &\textcolor{blue}{85.35$\pm$3.44} &\textcolor{blue}{84.57$\pm$3.49} &78.37$\pm$3.42 &\textcolor{blue}{74.71$\pm$2.91} &\textcolor{blue}{70.09$\pm$5.08} &87.03$\pm$2.74 &85.66$\pm$2.91 &\textcolor{blue}{84.66$\pm$3.41}	\\
				&OMVC	&75.35$\pm$2.11 &72.85$\pm$3.17 &70.61$\pm$2.77 &68.45$\pm$3.22 &65.44$\pm$2.89 &63.34$\pm$4.36 &80.89$\pm$3.05 &77.96$\pm$2.33 &74.52$\pm$3.55	\\									
				&OPIMC	&79.82$\pm$8.32	&74.57$\pm$3.81	&71.25$\pm$6.27	&69.92$\pm$6.36	&66.87$\pm$1.86	&64.65$\pm$6.94	&81.56$\pm$5.12	&79.02$\pm$1.27	&78.21$\pm$4.01\\
				&MKKM-IK-MKC	&74.34$\pm$0.34	&75.92$\pm$0.93	&71.22$\pm$1.19	&65.21$\pm$0.32	&66.83$\pm$1.24	&65.27$\pm$1.66	&78.96$\pm$0.07	&79.18$\pm$0.16	&79.94$\pm$1.03\\
				&PIC	&77.62$\pm$0.61 &76.45$\pm$8.52 &73.65$\pm$9.51 &78.10$\pm$4.58 &72.96$\pm$8.05 &68.46$\pm$10.41 &81.33$\pm$0.72 &77.83$\pm$0.77 &74.55$\pm$0.85	\\	
				&UEAF	&80.36$\pm$0.10 &71.22$\pm$0.68 &64.37$\pm$1.13 &67.11$\pm$0.52 &55.52$\pm$2.55 &47.97$\pm$1.50 &81.67$\pm$0.13 &73.32$\pm$0.70 &68.49$\pm$1.21	\\	
				&Ours	&\textcolor{red}{\textbf{90.45$\pm$3.57}}	&\textcolor{red}{\textbf{89.51$\pm$3.86}}	&\textcolor{red}{\textbf{89.11$\pm$1.40}}	&\textcolor{red}{\textbf{86.72$\pm$2.52}}	&\textcolor{red}{\textbf{85.57$\pm$3.76}} &\textcolor{red}{\textbf{84.73$\pm$2.97}}	&\textcolor{red}{\textbf{90.53$\pm$0.90}}	&\textcolor{red}{\textbf{89.60$\pm$0.90}}	&\textcolor{red}{\textbf{89.11$\pm$0.89}}   \\
				\hline
			\end{tabular}
		}
		\label{tab:res2}%
	\end{table*}
	
\begin{table*}[htbp]
	\small
	\centering
	\caption{METRICS OF TEN/TWELVE IMC MODELS ON THE \textbf{HANDWRITTEN} DATABASE WITH DIFFERENT INCOMPLETE RATIOS AND \textbf{ANIMAL} DATABASE WITH DIFFERENT RATIOS OF PAIRED SAMPLES. The 1st/2nd BEST RESULTS ARE MARKED IN \textcolor{red}{RED}/\textcolor{blue}{BLUE}.}\resizebox{0.95\textwidth}{!}{
		\begin{tabular}{|c|l|ccc|ccc|ccc|}
			\hline
			\multicolumn{2}{|c|}{} & \multicolumn{3}{c|}{ACC (\%)}      & \multicolumn{3}{c|}{NMI (\%)}      & \multicolumn{3}{c|}{Purity (\%)} \\
			\hline
			\multicolumn{1}{|c|}{Dataset} & Method\textbackslash{}Rate & 30\% & 50\% & 70\% & 30\% & 50\% & 70\%  & 30\% & 50\% & 70\% \\
			\hline
			\multirow{11}{*}{\begin{turn}{0}Handwritten\end{turn}}
			&BSV	&51.49$\pm$2.29	&38.24$\pm$2.25	&27.15$\pm$1.31	&47.01$\pm$1.71	&32.21$\pm$1.00	&19.48$\pm$0.69	&53.69$\pm$1.54	&39.54$\pm$2.04	&27.76$\pm$1.09\\
   			&Concat	&55.48$\pm$1.57	&42.19$\pm$0.99	&28.31$\pm$0.75	&51.66$\pm$0.99	&38.24$\pm$1.59	&23.50$\pm$0.95	&57.32$\pm$1.15	&44.21$\pm$0.98	&30.45$\pm$0.80\\
			&MIC	&73.29$\pm$3.41	&61.27$\pm$3.16	&41.34$\pm$2.69	&65.39$\pm$2.08	&52.95$\pm$1.33	&34.71$\pm$2.11	&74.31$\pm$3.15	&62.89$\pm$3.08	&43.25$\pm$2.86\\
	    	&DAIMC	&\textcolor{blue}{86.73$\pm$0.79}	&81.92$\pm$0.88	&60.44$\pm$6.87	&76.65$\pm$1.07	&68.77$\pm$0.99	&47.10$\pm$4.79	&\textcolor{blue}{86.73$\pm$0.79}	&81.92$\pm$0.88	&61.24$\pm$0.42\\
   			&OMVC	&55.00$\pm$5.06	&36.40$\pm$4.93	&29.80$\pm$4.63	&44.99$\pm$4.56	&35.16$\pm$4.62	&25.83$\pm$8.37	&55.89$\pm$4.72	&38.51$\pm$4.87	&31.95$\pm$5.22\\
   			&OPIMC	&76.45$\pm$5.15	&69.50$\pm$6.54	&56.66$\pm$10.06	&73.74$\pm$3.42	&66.57$\pm$4.18	&51.86$\pm$7.97	&78.96$\pm$3.37	&72.00$\pm$6.39	&58.16$\pm$10.35\\
			&MKKM-IK-MKC	&69.07$\pm$0.73	&66.08$\pm$3.25	&55.55$\pm$1.39	&65.42$\pm$0.61	&59.04$\pm$2.69	&47.36$\pm$1.78 &73.12$\pm$0.61	&66.58$\pm$3.26	&56.26$\pm$1.07\\
			&PIC	&83.90$\pm$0.17	&\textcolor{blue}{83.24$\pm$0.28}	&\textcolor{blue}{80.97$\pm$1.70}	&\textcolor{blue}{84.79$\pm$1.73}	&\textcolor{blue}{82.25$\pm$1.66}	&\textcolor{red}{\textbf{77.56$\pm$0.59}}	&85.92$\pm$1.12	&\textcolor{blue}{84.23$\pm$1.29}	&\textcolor{blue}{80.97$\pm$1.70}   \\
			&UEAF	&76.11$\pm$7.74	&65.39$\pm$5.09	&61.11$\pm$1.41	&69.37$\pm$3.31	&55.09$\pm$2.05	&50.56$\pm$1.11	&76.51$\pm$7.17	&66.49$\pm$4.18	&61.60$\pm$1.09\\
			&Ours	&\textcolor{red}{\textbf{93.63$\pm$0.88}}	&\textcolor{red}{\textbf{87.42$\pm$5.57}}	&\textcolor{red}{\textbf{85.75$\pm$0.85}}	&\textcolor{red}{\textbf{88.01$\pm$1.33}}	&\textcolor{red}{\textbf{82.75$\pm$2.37}}	&\textcolor{blue}{76.46$\pm$0.90}	&\textcolor{red}{\textbf{93.63$\pm$0.88}}	&\textcolor{red}{\textbf{88.10$\pm$4.64}}	&\textcolor{red}{\textbf{85.75$\pm$0.75}}\\
			
			\hline
			\multirow{8}{*}{\begin{turn}{0}Animal\end{turn}}
			&BSV	&42.05$\pm$1.20	&48.63$\pm$1.89	&56.22$\pm$1.20	&48.16$\pm$0.44	&55.91$\pm$0.58	&63.99$\pm$0.38	&45.20$\pm$0.88	&52.26$\pm$1.19	&60.31$\pm$0.78\\
			&Concat	&42.79$\pm$0.67	&49.34$\pm$1.39	&53.99$\pm$0.99	&55.46$\pm$0.16	&59.31$\pm$0.38	&63.88$\pm$0.35	&48.12$\pm$0.45	&53.24$\pm$0.88	&59.26$\pm$0.81\\
			&MIC	&43.38$\pm$0.63	&45.88$\pm$0.34	&49.15$\pm$0.88	&52.79$\pm$0.77	&55.69$\pm$0.36	&59.30$\pm$0.54	&49.21$\pm$0.78	&52.31$\pm$0.34	&55.33$\pm$0.64\\
			&DAIMC	&50.18$\pm$2.18	&53.87$\pm$1.36	&56.42$\pm$1.37	&55.03$\pm$1.03	&59.36$\pm$1.16	&62.76$\pm$0.46	&54.82$\pm$1.57	&59.51$\pm$1.65	&62.12$\pm$1.04\\
			&OMVC	&42.51$\pm$0.89	&43.98$\pm$0.77	&46.39$\pm$1.02	&50.77$\pm$0.63	&53.11$\pm$0.83	&55.38$\pm$0.46	&47.33$\pm$0.66	&50.42$\pm$0.91	&52.97$\pm$0.76\\
			&OPIMC	&46.33$\pm$2.14	&53.14$\pm$1.38	&53.88$\pm$1.26	&52.34$\pm$0.69	&58.51$\pm$0.46	&62.04$\pm$0.26	&49.49$\pm$1.41	&56.23$\pm$1.20	&57.91$\pm$0.43\\
			&MKKM-IK-MKC	&51.77$\pm$0.48	&\textcolor{blue}{57.75$\pm$0.38}	&\textcolor{blue}{61.18$\pm$0.59}	&56.54$\pm$0.33	&61.66$\pm$0.22	&\textcolor{blue}{66.28$\pm$0.27} &56.14$\pm$0.48	&62.14$\pm$0.41	&\textcolor{blue}{66.40$\pm$0.53}\\
			&PIC	&\textcolor{blue}{55.94$\pm$0.78}	&56.84$\pm$1.55	&57.67$\pm$1.03	&\textcolor{blue}{62.35$\pm$0.46}	&\textcolor{blue}{64.37$\pm$0.64}	&65.82$\pm$0.26 &\textcolor{red}{\textbf{63.07$\pm$0.44}}	&\textcolor{blue}{64.75$\pm$1.57}	&65.42$\pm$0.38\\
			&UEAF	&45.73$\pm$12.9	&51.86$\pm$6.48	&58.19$\pm$3.04	&51.61$\pm$12.87	&58.43$\pm$7.53	&64.92$\pm$3.95 &49.10$\pm$0.27	&55.36$\pm$0.36	&63.02$\pm$0.47\\
			&COMPLETER	&28.44$\pm$2.88	&29.28$\pm$4.45	&34.09$\pm$2.94	&46.48$\pm$2.00	&47.01$\pm$3.92	&50.43$\pm$1.45 &31.42$\pm$2.94	&31.23$\pm$5.26	&36.13$\pm$2.61\\
			&SURE	&27.33$\pm$0.21	&30.97$\pm$0.12	&36.28$\pm$0.10	&42.88$\pm$0.12	&50.20$\pm$0.14	&53.12$\pm$0.17 &29.09$\pm$0.26	&35.04$\pm$0.21	&39.22$\pm$0.13\\
			&Ours	&\textcolor{red}{\textbf{56.30$\pm$0.53}}	&\textcolor{red}{\textbf{60.68$\pm$0.81}}	&\textcolor{red}{\textbf{64.83$\pm$1.11}}	&\textcolor{red}{\textbf{62.64$\pm$0.30}}	&\textcolor{red}{\textbf{66.79$\pm$0.38}}	&\textcolor{red}{\textbf{71.21$\pm$0.57}}	&\textcolor{blue}{60.14$\pm$0.43}	&\textcolor{red}{\textbf{64.92$\pm$0.48}}	&\textcolor{red}{\textbf{69.17$\pm$0.54}}\\
			\hline
	\end{tabular}}
	\label{tab:res3}
\end{table*}

	Table \ref{tab:res1}-Table \ref{tab:res3} list the experimental results on different incomplete datasets, including the mean and standard deviation value of ACC (\%), NMI (\%), and purity (\%). We can further obtain the following observations and analysis from the tables.
	
	(1) Obviously, by comparing the state-of-the-art methods listed in tables, our proposed method almost achieves the best performance. For example, on the widely used Handwritten database with a 30\% missing rate, the proposed method achieves approximately 10\%, 3\%, and 8\% performance improvements over the second-best method (\textit{i.e.}, PIC) in terms of ACC, NMI, and purity, respectively. In addition, on the large-scale Animal dataset with 70\% paired samples, the proposed method is also far superior to the second place in the three measures, \textit{i.e.}, 8.41\% ACC over DAIMC, 7.22\% NMI over BSV, and 7.05\% purity over DAIMC.
	
	(2) In addition to our LSIMVC, PIC and UEAF also exhibit competitive performance on different databases. The PIC and UEAF exploit local information of data to varying degrees in the process of consensus representation learning. Specifically, PIC transforms the problem from feature missing into similarity information missing of graph space; UEAF utilizes the non-negative similarity graph which preserves the similarity of any two samples to induce low-dimensional representation learning. Therefore, we can confirm that taking full advantage of the internal structure of the data plays a positive role in improving the performance of IMC.
	
	(3) It can be clearly observed from tables that the experimental results of all methods have the same law, that is, for the Animal database, all three indicators' values increase with the increase of the rate of paired samples, and for other databases with randomly removed samples, the indicators show a downward trend as the view missing rate increases. This phenomenon demonstrates that it becomes more and more difficult to learn and represent the consistent information shared among multi-view as the view missing rate increases.
	
	(4) Both deep methods (COMPLETER and SURE) show unsatisfactory performance on the Animal dataset, especially on the ACC metric, although we experiment with the open-source codes of these two methods and fine-tune them based on the recommended parameters (For SURE, the result we obtained is better than that in \cite{yang2022robust}). We consider the following two possible reasons: On the one hand, since both views of the Animal dataset are the high-level features extracted from the deep neural network, overfitting occurs on the two deep methods using autoencoders; on the other hand, the contrastive learning strategy adopted by these two methods destroys the original nearest neighbor structure of the dataset.
	
\begin{figure*}[t]
		\centering
		\vspace{-0.8cm}
		\subfloat[BSV]{
			\label{fig:tsne1}
			\includegraphics[width=1.35in,height=1.1in]{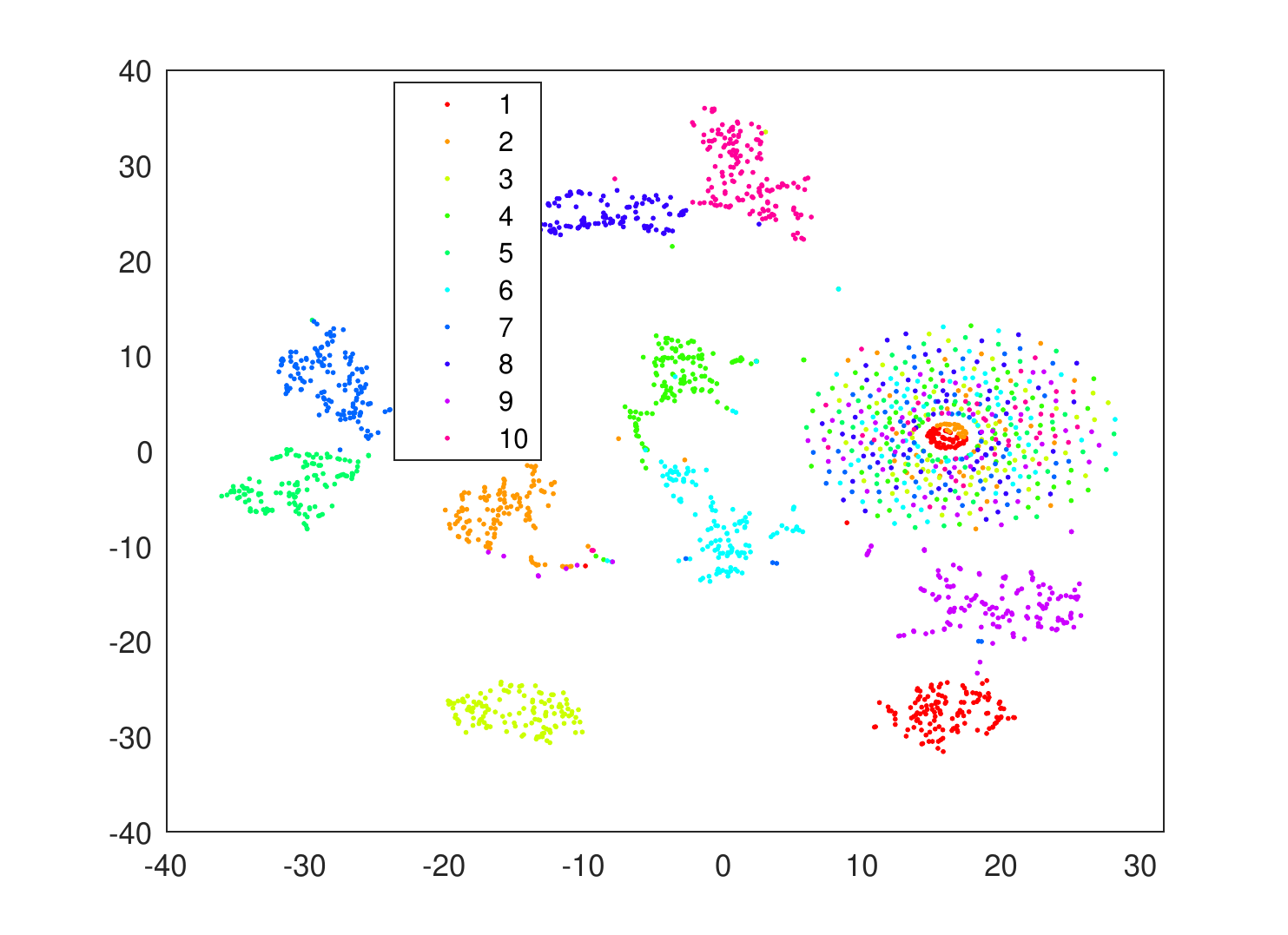}
		}
		\subfloat[Concat]{
			\label{fig:tsne2}
			\includegraphics[width=1.35in,height=1.1in]{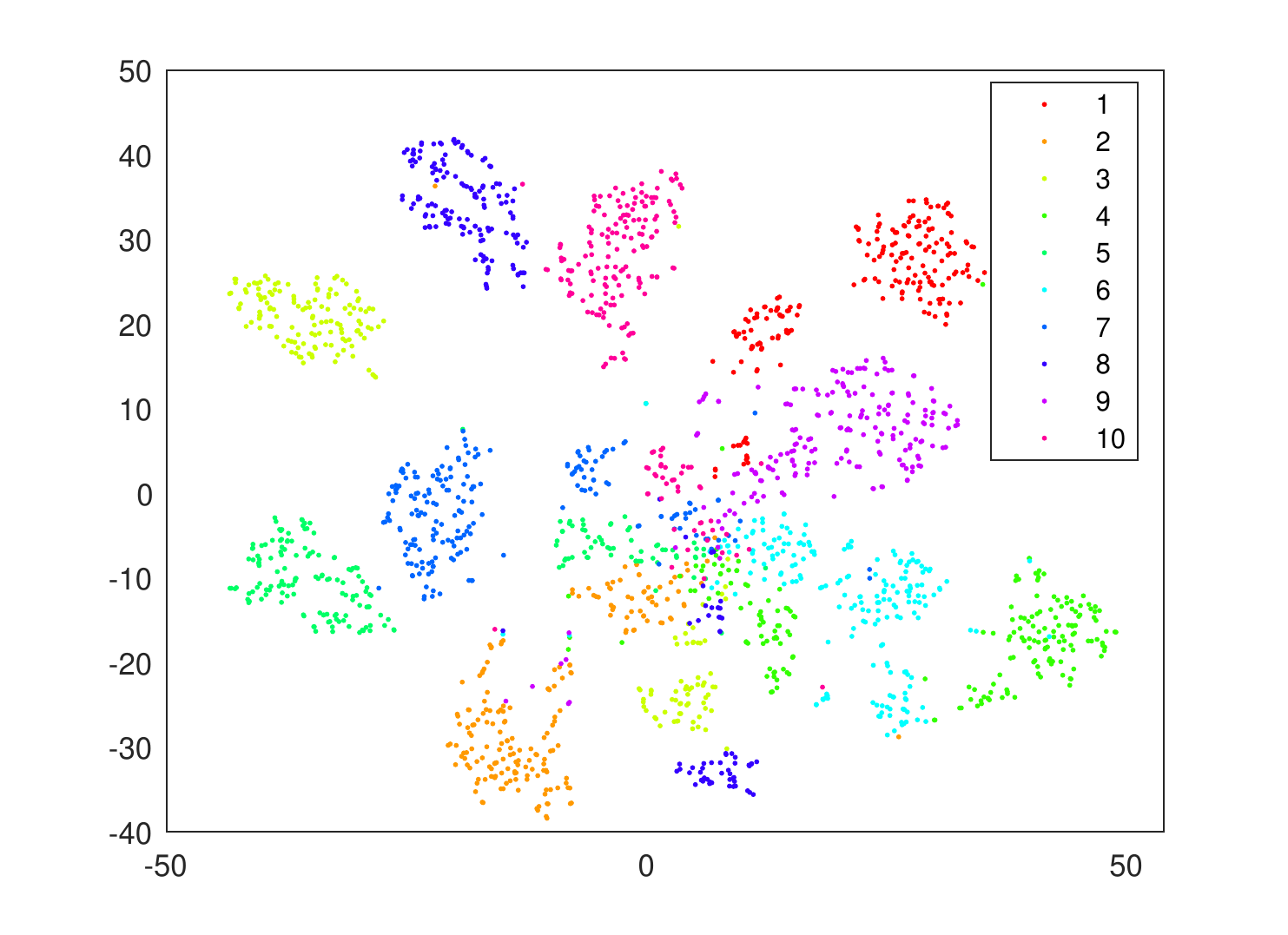}
		}
		\subfloat[MIC]{
			\label{fig:tsne3}
			\includegraphics[width=1.35in,height=1.1in]{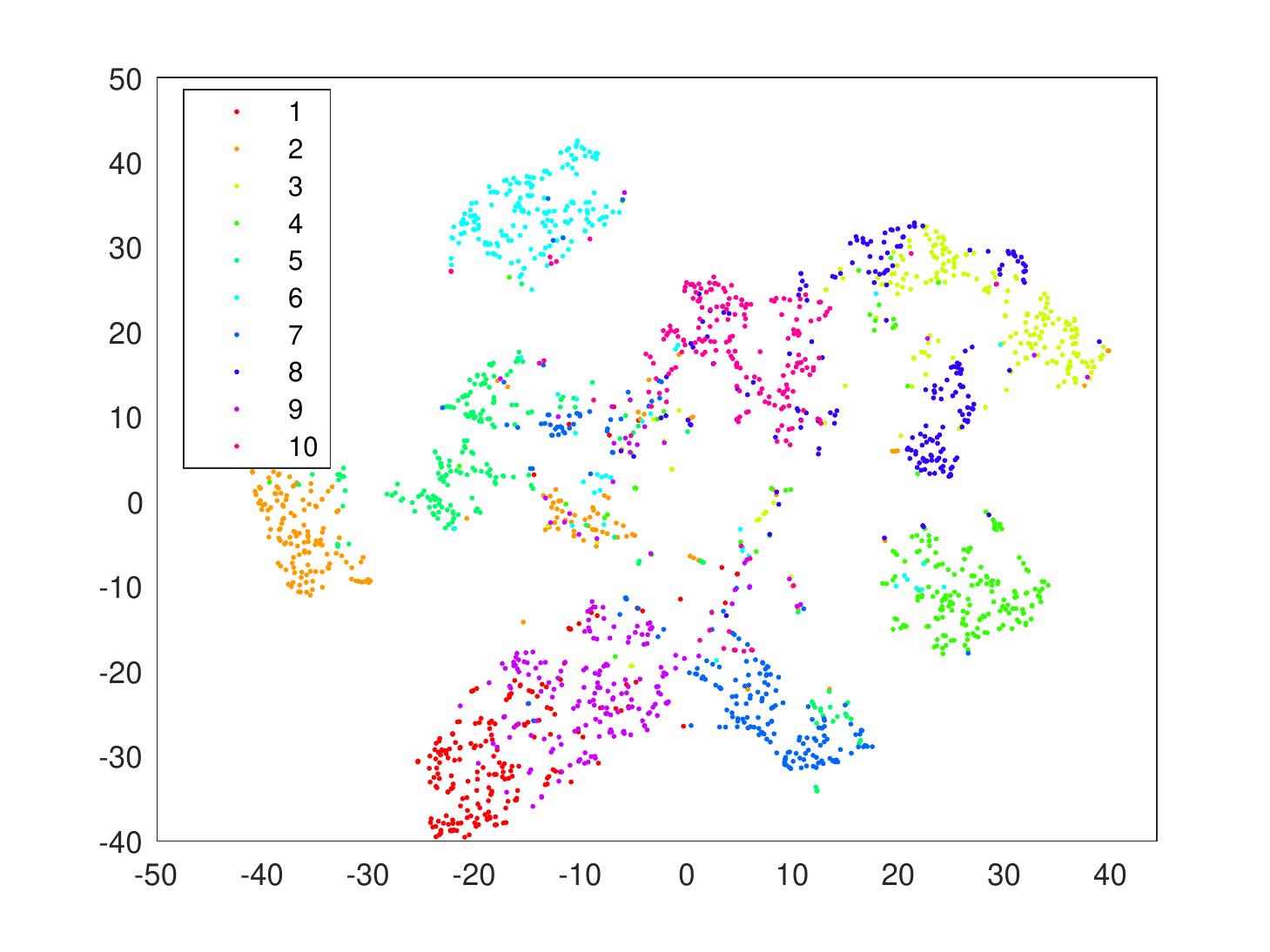}
		}
		\subfloat[DAIMC]{
			\label{fig:tsne4}
			\includegraphics[width=1.35in,height=1.1in]{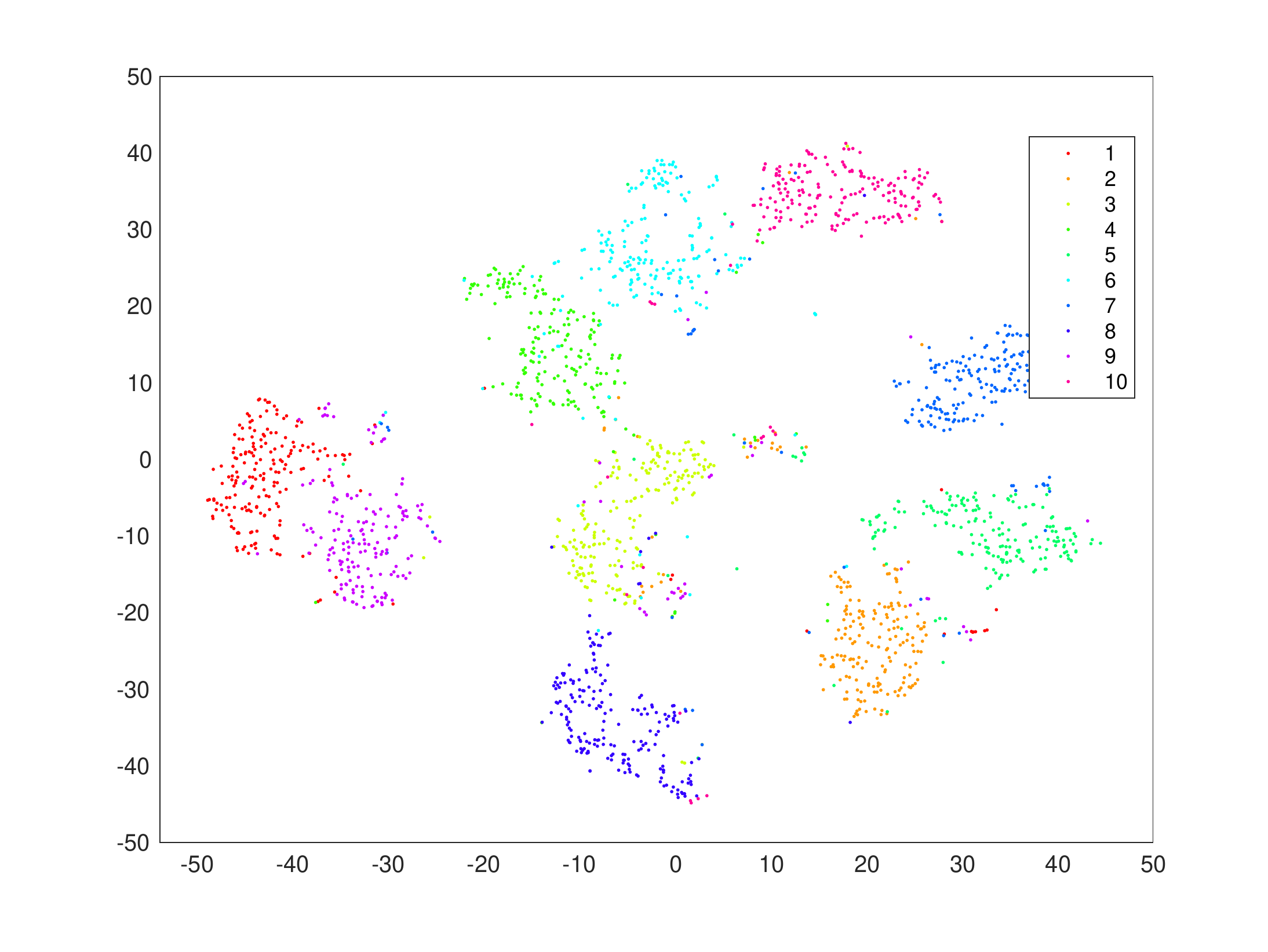}
		}
		\subfloat[OMVC]{
			\label{fig:tsne5}
			\includegraphics[width=1.35in,height=1.1in]{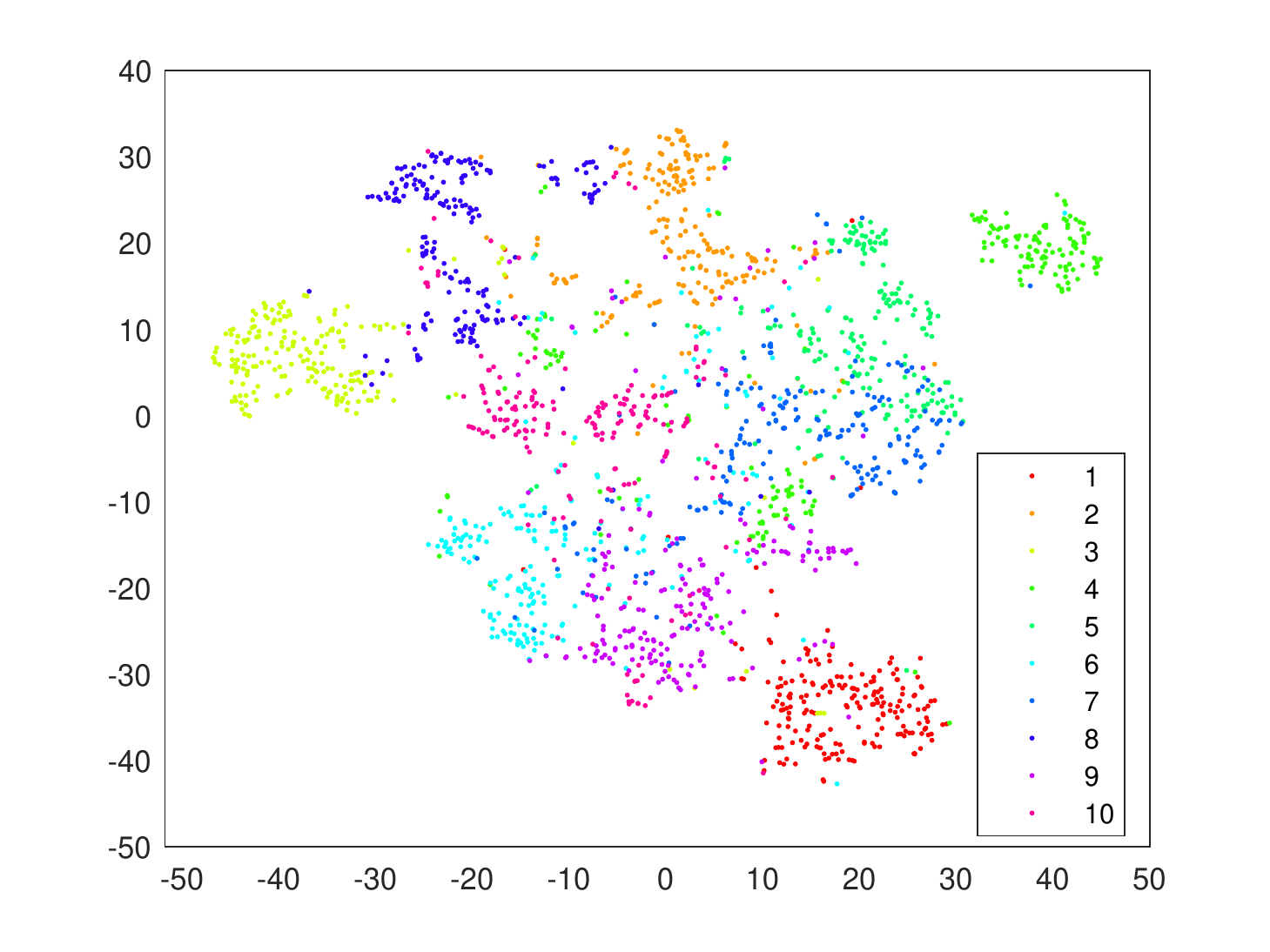}
		}
		
		\subfloat[MKKM-IK-MKC]{
			\label{fig:tsne7}
			\includegraphics[width=1.35in,height=1.1in]{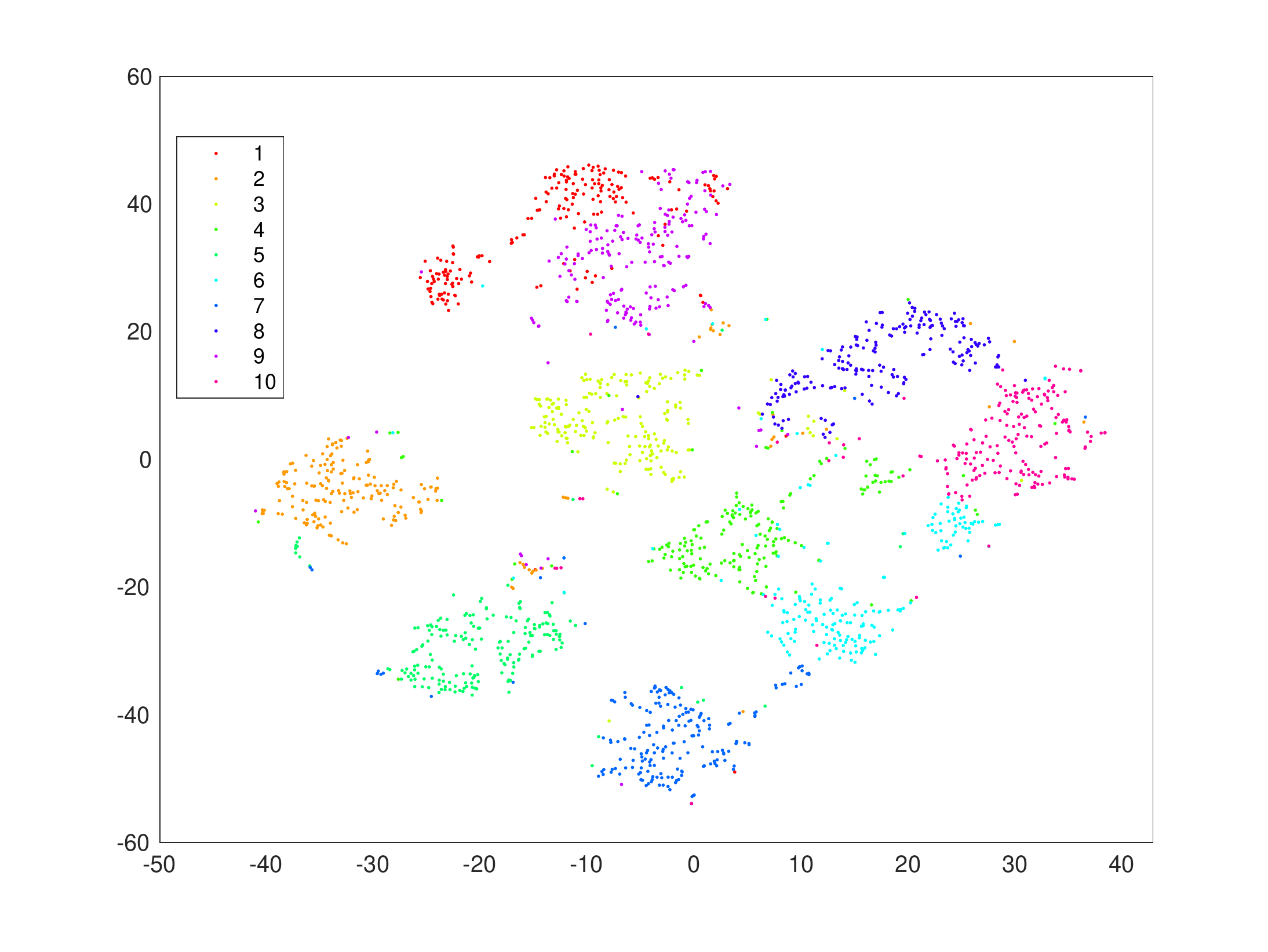}
		}
		\subfloat[PIC]{
			\label{fig:tsne8}
			\includegraphics[width=1.35in,height=1.1in]{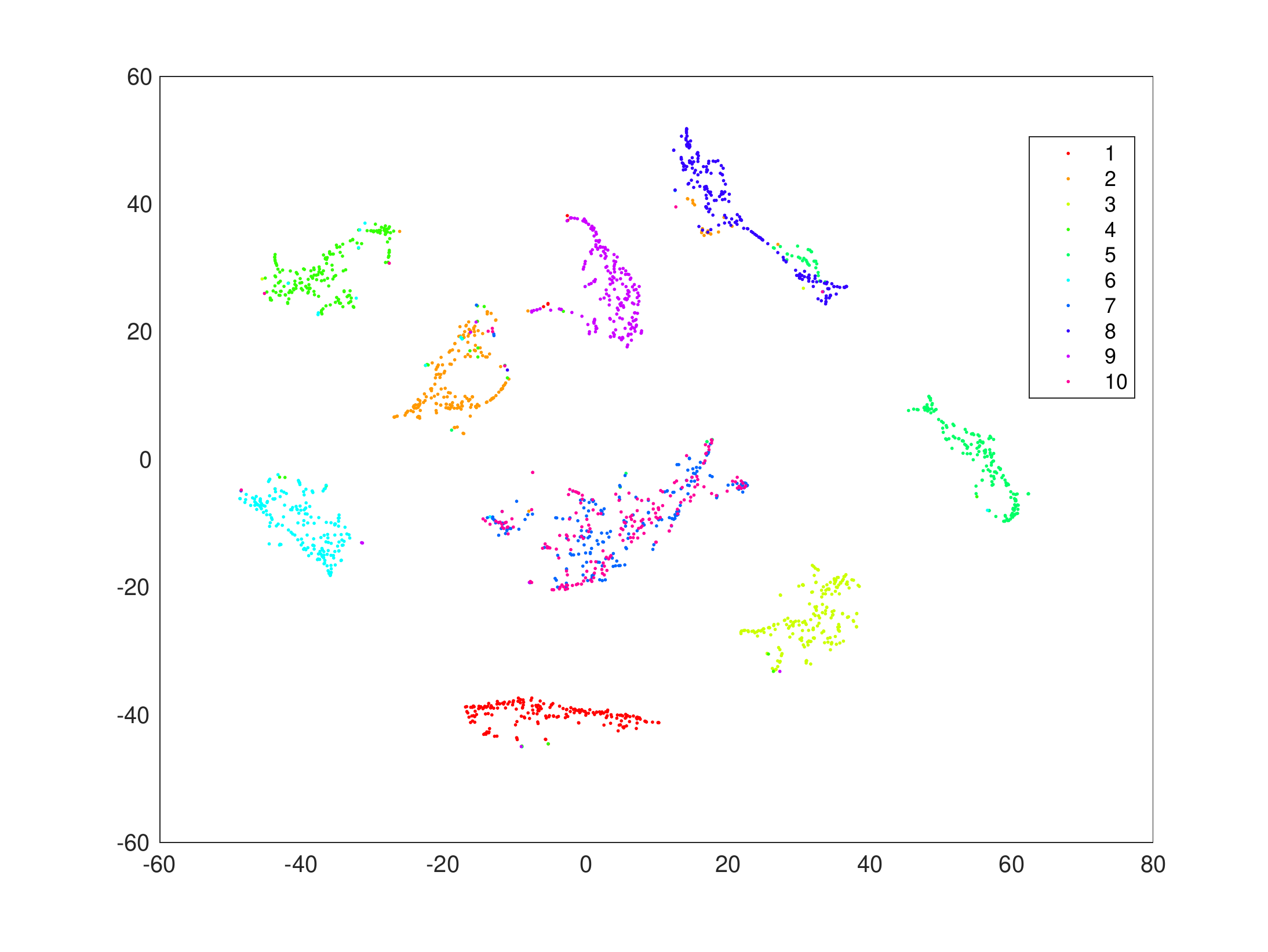}
		}
		\subfloat[UEAF]{
			\label{fig:tsne9}
			\includegraphics[width=1.35in,height=1.1in]{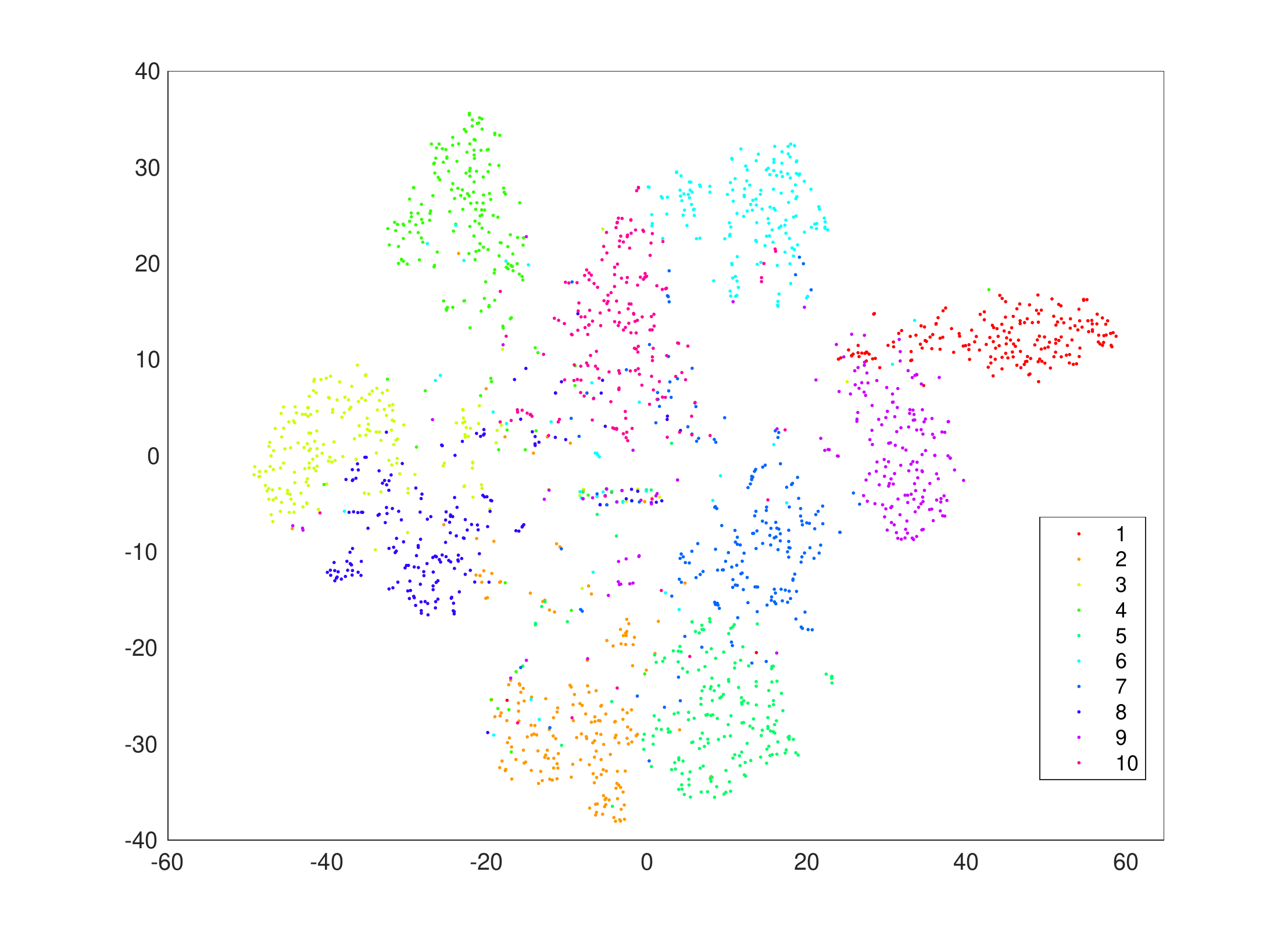}
		}
		\subfloat[LSIMVC]{
			\label{fig:tsne10}
			\includegraphics[width=1.35in,height=1.1in]{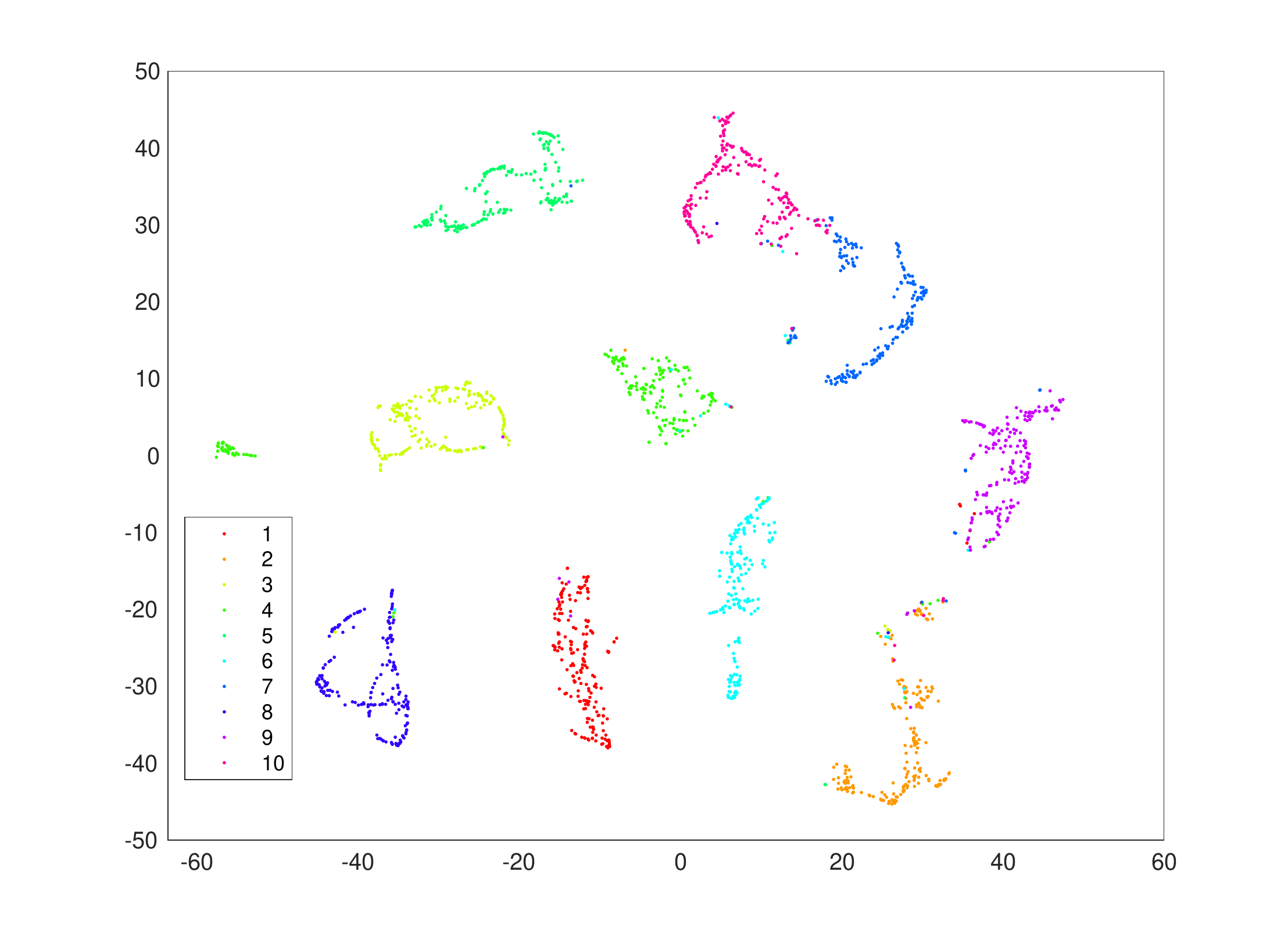}
		}
		\caption{Feature space visualization of final clustering representation of different methods via t-SNE on the Handwritten database with a 30\% incomplete rate.}
		\label{fig:tsne}
	\end{figure*}

	As shown in Fig. \ref{fig:tsne}, in order to observe their clustering performance more intuitively, we represent the final clustering representation of different approaches via t-SNE \cite{van2008visualizing} on the Handwritten database when 30\% of instances are removed. Obviously, from the distribution of the samples, the representation obtained by our method shows better intra-class compactness and inter-class separation, which benefits from our method's ability to preserve the original structure of the data. It should be noted that we do not plot the representation of OPIMC in the figure because it directly produces the one-hot label matrix instead of the feature matrix input to k-means. 
	\subsection{Parameter sensitivity analysis}
	\begin{figure}[t!]
		\centering
		
		\subfloat[BBCSport]{
			\label{fig:bbc2}
			\includegraphics[width=0.5\linewidth]{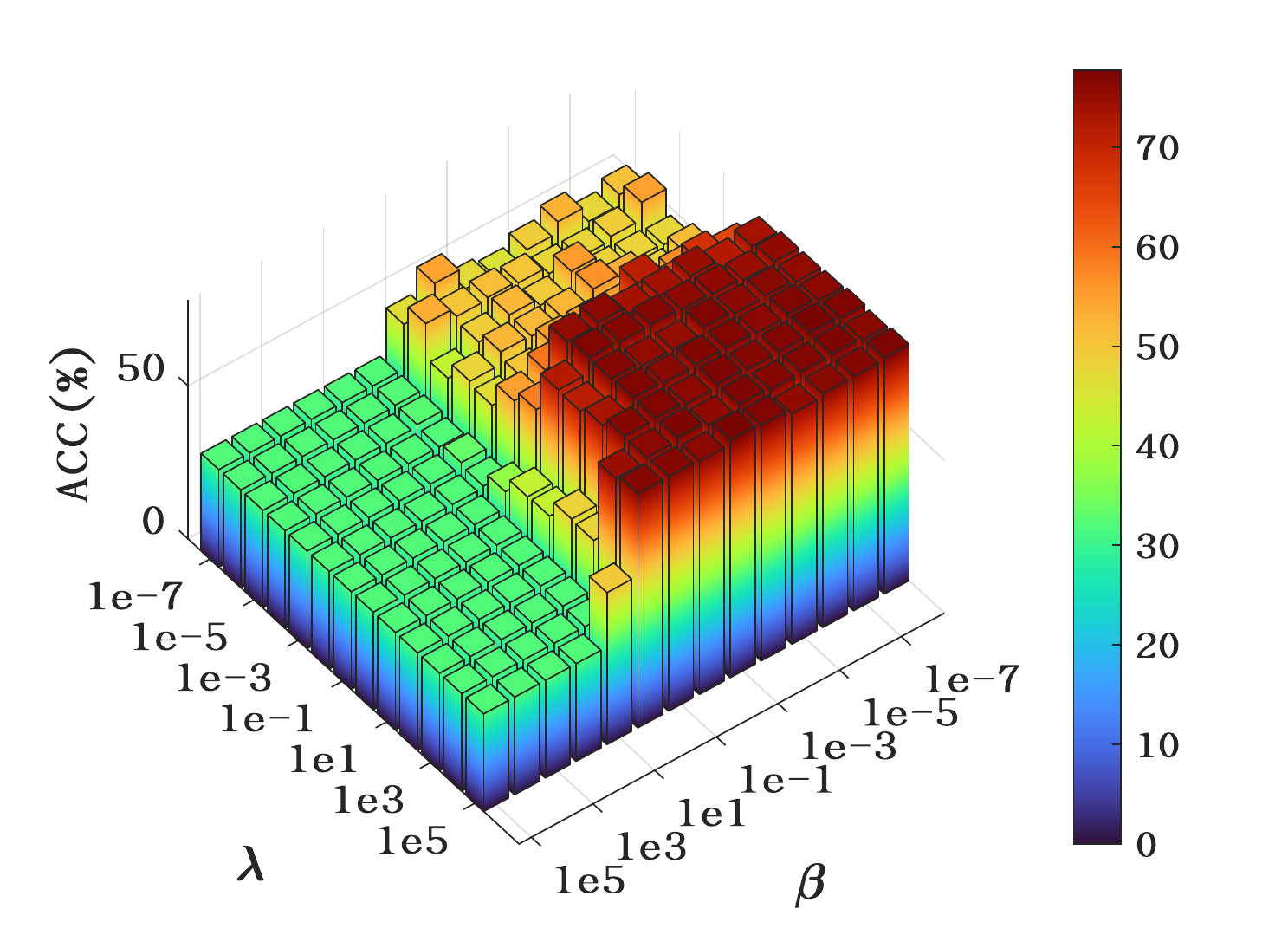}
		}
		\subfloat[Caltech7]{
			\label{fig:C72}
			\includegraphics[width=0.5\linewidth]{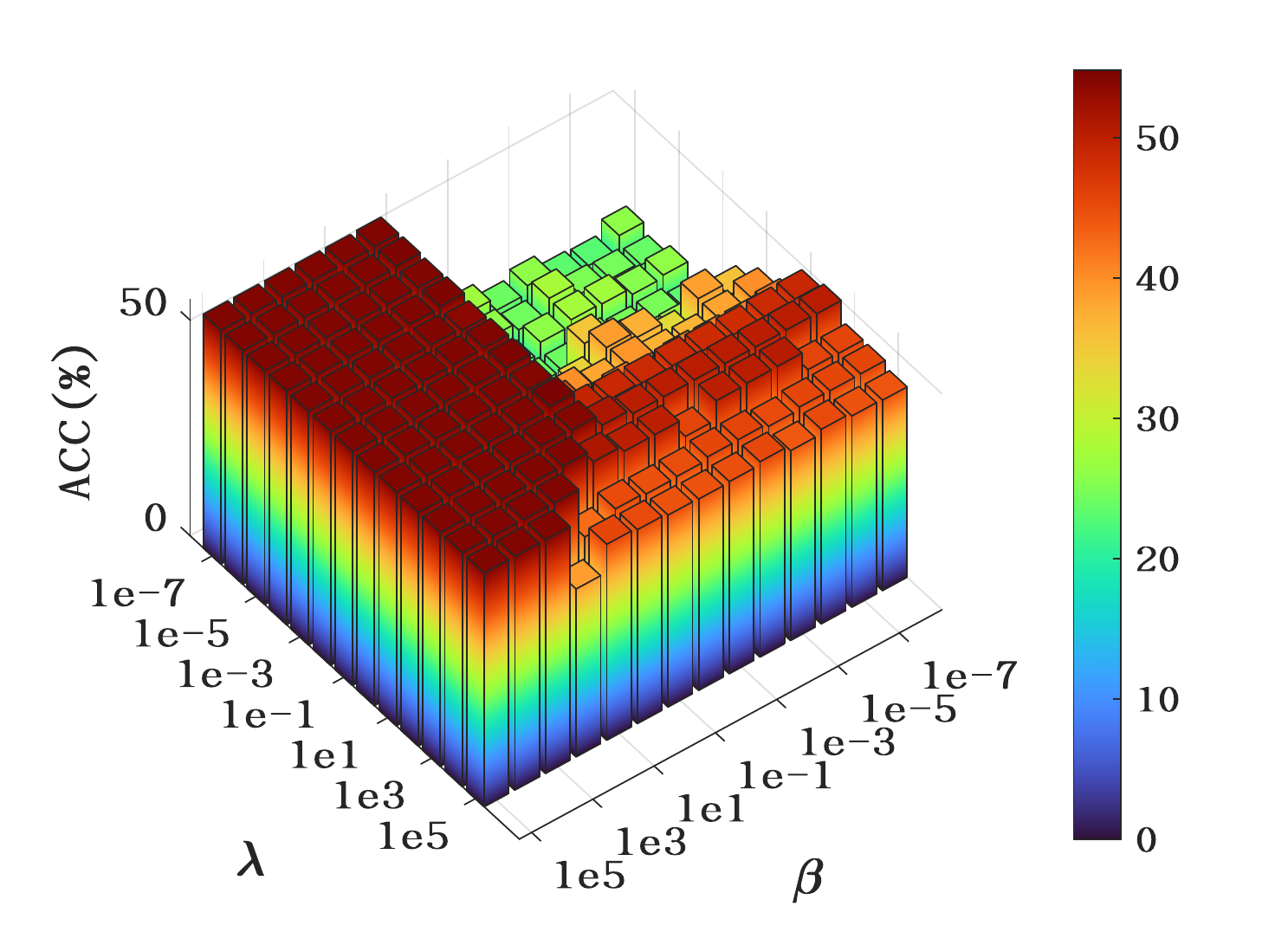}
		}
		
		\caption{ACC (\%) vs. candidate sets for the parameters $\lambda$ and $\beta$ of the LSIMVC on the (a) BBCSport database with a 30\% incomplete rate and (b) Caltech7 database with a 50\% incomplete rate.}
		\label{fig:lam_beta}
	\end{figure}
	
	\begin{figure}[htbp]
		\centering
		\subfloat[BBCSport]{
			\label{fig:bbc1}
			\includegraphics[width=0.5\linewidth]{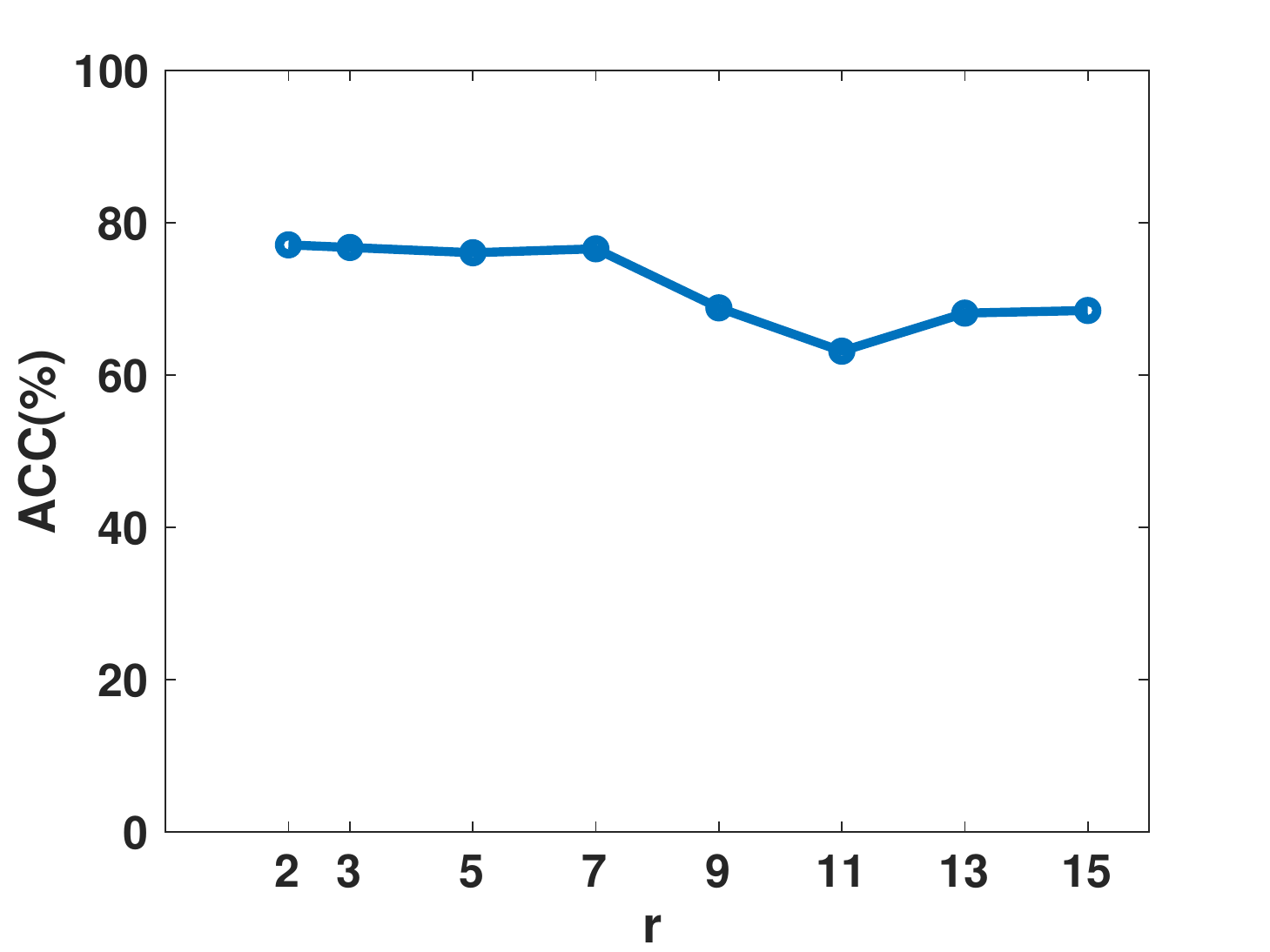}
		}
		\subfloat[Caltech7]{
			\label{fig:C71}
			\includegraphics[width=0.5\linewidth]{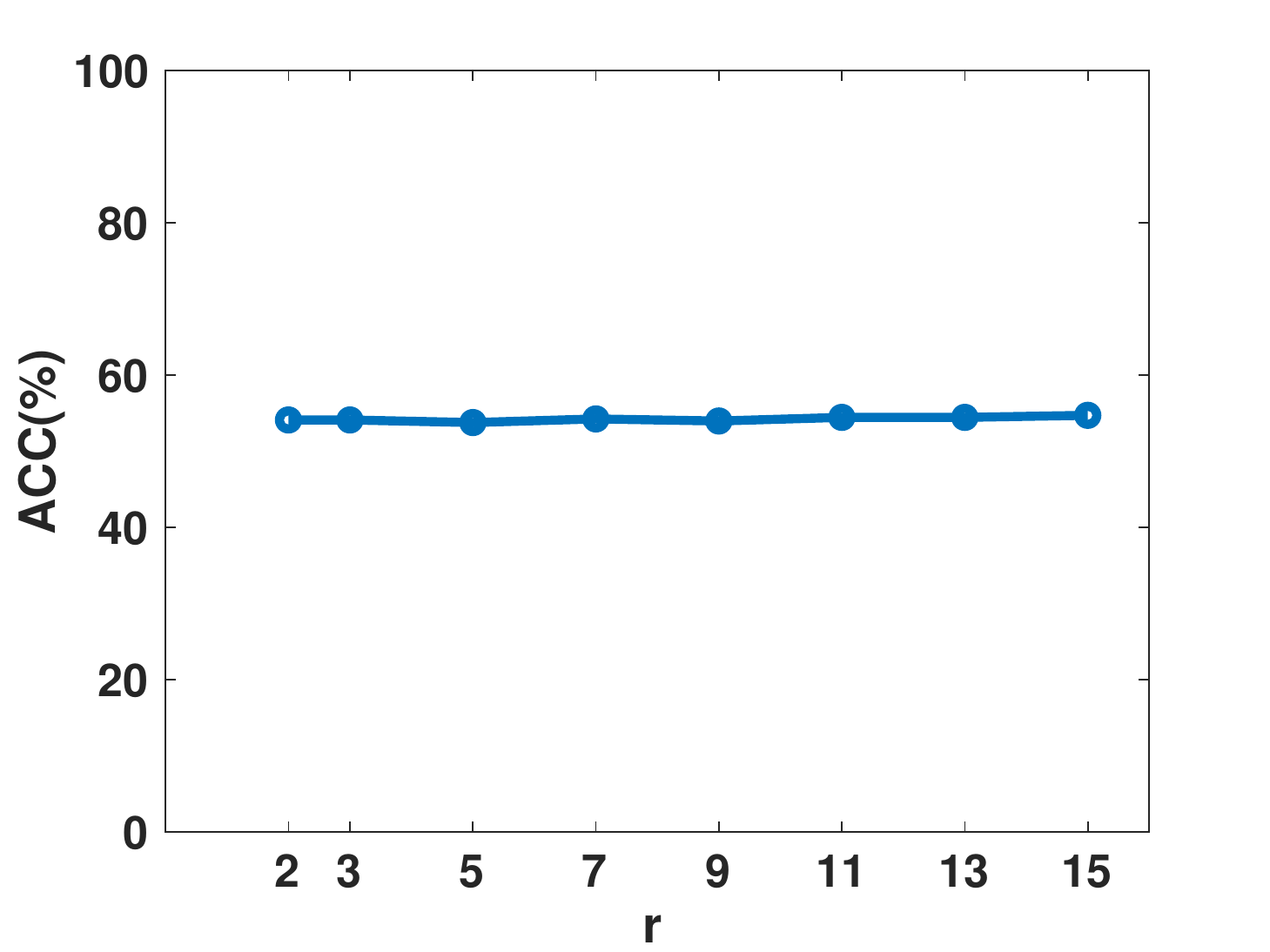}
		}
		
		\caption{ACC (\%) vs. parameter $r$ of the  LSIMVC on the (a) BBCSport database with a 30\% incomplete rate and (b) Caltech7 database with a 50\% incomplete rate.}
		\label{fig:r}
	\end{figure}
	\begin{table*}[htbp]
			\centering
			\caption{ABLATION EXPERIMENTS ON THE BBCSPORT AND 3SOURCES DATABASES WITH DIFFERENT INCOMPLETE RATIOS.}
			\resizebox{0.95\textwidth}{!}{
				\begin{tabular}{|c|l|ccc|ccc|ccc|}
					\hline
					\multicolumn{2}{|c|}{} & \multicolumn{3}{c|}{ACC (\%)}      & \multicolumn{3}{c|}{NMI (\%)}      & \multicolumn{3}{c|}{Purity (\%)} \\
					\hline
					\multicolumn{1}{|c|}{Dataset} & Method\textbackslash{}Rate & 10\% & 30\% & 50\% &10\% & 30\% & 50\% &10\% & 30\% & 50\% \\
					\hline	
					\multirow{4}{*}{\begin{turn}{0}BBCSport\end{turn}}
					&LSIMVC \textit{\textbf{w/o}} Weight	&77.93$\pm$1.31 &75.17$\pm$3.01 &66.72$\pm$8.21 &67.75$\pm$2.72 &67.54$\pm$5.10 &56.52$\pm$8.45 &87.07$\pm$0.72 &86.38$\pm$0.71 &77.59$\pm$0.59\\
					&LSIMVC \textit{\textbf{w/o}} Sparsity	&77.76$\pm$1.66 &76.90$\pm$4.66 &68.28$\pm$11.10 &68.16$\pm$1.17 &66.63$\pm$5.30 &56.64$\pm$7.26 &87.76$\pm$0.73 &85.86$\pm$0.70 &77.59$\pm$0.60\\

					&LSIMVC \textit{\textbf{w/o}} Graph	&62.07$\pm$4.99 &56.72$\pm$5.53 &51.90$\pm$8.52 &44.85$\pm$7.30 &36.59$\pm$8.41 &33.43$\pm$12.07 &69.83$\pm$0.49 &63.45$\pm$0.44 &60.86$\pm$0.41\\
					&LSIMVC	&\textbf{80.69$\pm$3.69}	&\textbf{78.79$\pm$1.68}	&\textbf{72.07$\pm$2.25}	&\textbf{73.57$\pm$1.66}	&\textbf{67.71$\pm$4.02}	&\textbf{58.65$\pm$2.33}	&\textbf{89.83$\pm$1.12}	&\textbf{87.76$\pm$0.73}	&\textbf{80.17$\pm$2.20}\\
					\hline
					\multirow{4}{*}{\begin{turn}{0}3Sources\end{turn}}
					&LSIMVC \textit{\textbf{w/o}} Weight	&72.43$\pm$1.30 &71.72$\pm$4.55 &65.09$\pm$2.99 &63.87$\pm$1.33 &63.15$\pm$2.11 &56.84$\pm$3.31 &79.76$\pm$0.68 &79.53$\pm$0.65 &76.80$\pm$0.61\\
					&LSIMVC \textit{\textbf{w/o}} Sparsity	&71.72$\pm$1.64 &70.18$\pm$3.23 &63.67$\pm$3.34 &63.39$\pm$1.60 &63.62$\pm$3.80 &57.35$\pm$2.20 &79.17$\pm$0.68 &77.40$\pm$0.65 &76.09$\pm$0.63\\

					&LSIMVC \textit{\textbf{w/o}} Graph	&62.96$\pm$1.90 &58.82$\pm$3.52 &57.99$\pm$2.81 &60.27$\pm$2.23 &50.03$\pm$5.39 &53.93$\pm$3.09 &77.04$\pm$0.61 &71.24$\pm$0.56 &73.96$\pm$0.54\\
					&LSIMVC	&\textbf{74.79$\pm$4.84}	&\textbf{74.32$\pm$4.54}	&\textbf{67.57$\pm$4.17}	&\textbf{67.87$\pm$3.44}	&\textbf{66.16$\pm$4.05}	&\textbf{59.01$\pm$3.67}	&\textbf{83.31$\pm$2.76}	&\textbf{80.83$\pm$3.66}	&\textbf{76.33$\pm$3.05}\\
					\hline
				\end{tabular}
			}
			\label{tab:abl}
		\end{table*}
	In addition to the parameter $\gamma$, which is set to 1, the proposed LSIMVC contains three adjustable parameters, \textit{i.e.}, $\lambda$, $\beta$, and $r$, where $\lambda$ and $\beta$ are penalty parameters, and $r$ is a smoothing parameter. In order to obtain the optimal parameters, we conduct the grid parameters search approach on each database\cite{wen2018inter}. In this subsection, we analyze the sensitivity of ACC \textit{w.r.t.} these three parameters on two incomplete multi-view datasets, \textit{i.e.}, the BBCSport with a 30\% missing rate and Caltech7 with a  50\% missing rate.
	
	For parameters $\lambda$ and $\beta$, we define candidate sets $\left\{ 10^{x} | -7 \le x \le 5, x \in Z \right\}$ and $\left\{ 10^{x} | -8 \le x \le 5, x \in Z \right\}$, respectively. Fig. \ref{fig:lam_beta} shows the relationship between ACC (\%) and the two penalty parameters on the two databases by fixing parameter $r$. For instances, on the BBCSport database, we can obtain relatively optimal clustering results when selecting parameters $\lambda \in [10^{-1}, 10^{5}]$ and $\beta \in [10^{-8}, 10^{-2}]$, respectively. For the Caltech7 database, when parameters $\lambda$ and $\beta$ locate in the range of $[10^{-7},10^{2}]$ and $ \in [1,10^{5}]$, respectively, it is easier for us to obtain a satisfied clustering performance. The above experiments show that the proposed LISMVC is insensitive to the penalty parameters $\lambda$ and $\beta$ to some extent.
	
	For smoothing parameter $r$, we select parameter $r$ from the candidate set $\{2, 3, 5, 7, 9, 11, 13, 15\}$ and fix the other penalty parameters.  Fig. \ref{fig:r} reports the changing trend of ACC with different parameters $r$ on the BBCSport database and the Caltech7 database with 30\% and 50\% absence rates, respectively. Obviously, our LISMVC can achieve a satisfactory clustering result on the BBCSport database if we set parameter $r$ from $[2, 7]$. And on the Caltech7 database, the selection of parameter $r$ can hardly cause fluctuations of ACC.
	
	Since there are significant differences in not only the description objects but also the feature extraction approaches, the optimal parameter combination of the same model presents diversity distribution on different databases. For example, it is not appropriate to impose a relatively large smoothing parameter when the database has views that are equally discriminating. To the best of our knowledge, How to adaptively select the most appropriate parameters according to the characteristics of different datasets to obtain the best clustering results is still worth exploring. Typically, we search for the optimal value of each parameter in its predefined parameter spaces by fixing the other parameters in the combination.
	
	\subsection{Ablation experiments}
	To verify the effectiveness of each component of our LSIMVC, we conduct ablation experiments on the BBCSport and 3Sources databases. On the basis of (\ref{eq.over}), we respectively remove the adaptive weighted coefficient $\alpha_{v}^{r}$, the sparse constraint term $\beta\left\|P^{(v)}\right\|_{1} $, and the graph embedding constraint by simply setting $W_{i,j}^{(v)}$ to construct the according degraded methods, \textit{i.e.}, 'LSIMVC \textit{\textbf{w/o}} Weight', 'LSIMVC \textit{\textbf{w/o}} Sparsity', and 'LSIMVC \textit{\textbf{w/o}} Graph'. As shown in Table \ref{tab:abl}, all three degraded methods achieve lower performance than the complete LSIMVC. And it is easy to find that the performance of the LSIMVC without graph embedding constraint is greatly lower than that of the complete method, which indicates that learning the local structure within view can significantly improve the clustering performance. Overall, above phenomena fully verify the effectiveness of the three components in our LSIMVC.
	
	\subsection{Convergence analysis}
	\textbf{Theoretical}: Recalling the optimization steps in Section \uppercase\expandafter{\romannumeral3}, we divide the optimization problem of the target model (\ref{eq.over}) with variables $\left\{\left\{U^{(v)},P^{(v)}\right\}_{v=1}^l,\alpha,Q\right\}$ into four sub-problems, and iteratively optimize $U^{(v)}$, $P^{(v)}$, $Q$, and $\alpha$ in turn. For each optimization subprocess, we optimize the objective function with a single variable so that the value of the objective function is monotonically non-increasing. Therefore, during the complete iterative optimization process, the value of the objective function is monotonically non-increasing with 0 as the lower bound.
	
\begin{figure}[t]
		
		\centering
		\subfloat[BBCSport]{
			\label{fig.conva}
			\includegraphics[width=0.48\linewidth]{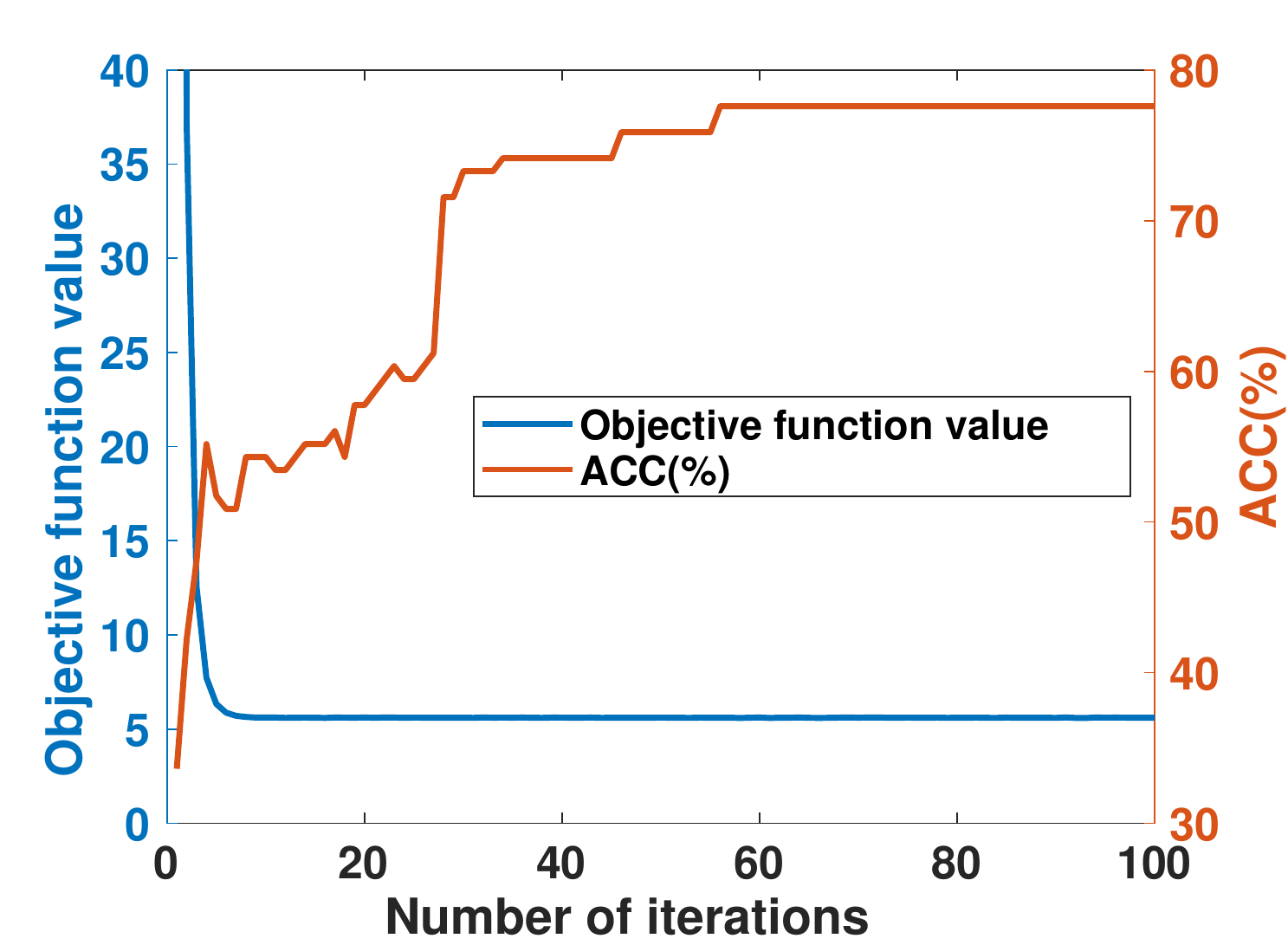}
		}
		\subfloat[Caltech7]{
			\label{fig.convb}
			\includegraphics[width=0.48\linewidth]{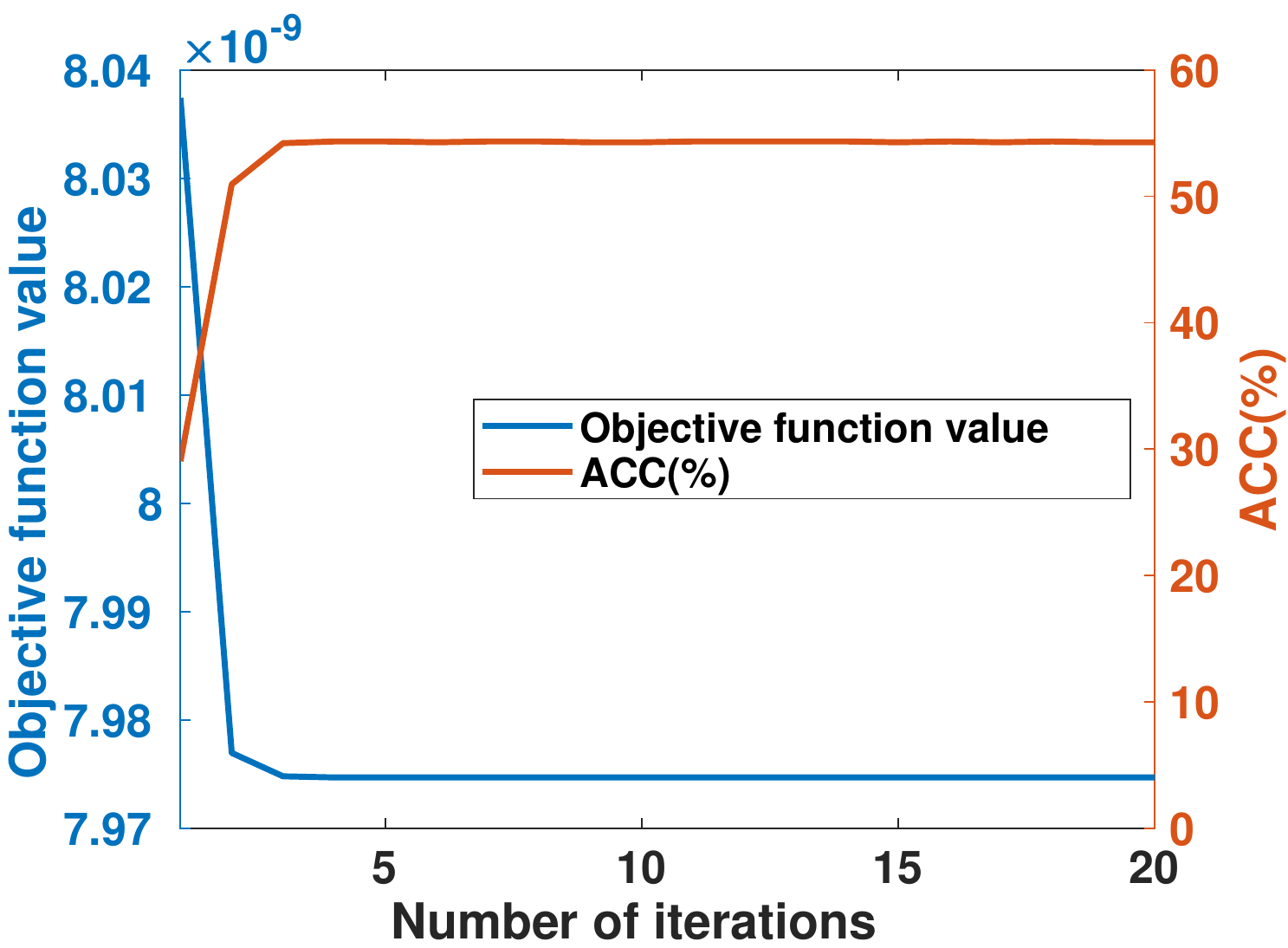}
		}
		\caption{The objective function value and clustering accuracy versus the number of iterations of the proposed method on the (a) BBCSport dataset where the missing-view rate is set as 30\% and the (b) Caltech7 dataset where the missing-view rate is set as 50\%.}
		\label{fig:objacc}

	\end{figure}

	\textbf{Experimental}: Fig. \ref{fig:objacc} shows the changing trend of the objective function value and ACC with the number of iterations on the BBCSport database and Caltech7 database where the missing-view rates are set as 30\% and 50\%, respectively. From Fig. \ref{fig:objacc}, we can observe that as the number of iterations increases, the objective function value decreases rapidly, and the ACC presents a significant upward trend and quickly reaches a stable level. It is worth noting that in Fig. \ref{fig:objacc}(a), the upward trend of the ACC curve is not strictly monotonically increasing. This is because in the step-by-step optimization process, the variables in the early stage of optimization change greatly, so the clustering results thus obtained by k-means also fluctuate accordingly. Overall, the results show that our proposed method has a good convergence property.
	
	\section{Conclusion}
	\label{chap:Cond}
	In this paper, a novel IMC framework named LSIMVC is proposed. Different from existing methods, the proposed method comprehensively takes into account the sparse representations of individual views, the consensus representation with localized structured information, and the discriminating abilities of different views. Importantly, by dexterously integrating the fusion graph embedding constraint into the consensus representation learning, our method can preserve the localized structure within the views in the common low-dimensional representation, which is of great benefit to acquire a discriminative common representation for clustering. In addition, incomplete multiple views, with different contributions to the consistent learning result, are adaptively weighted to match the corresponding discriminability during joint learning. These simple but influential approaches greatly enhance the performance and generalization capability of our method. Extensive experimental results conducted on six incomplete multi-view databases illustrate that the performance of our method exceeds the compared state-of-the-art IMC methods.
	
	In the future, we consider expanding the proposed core idea to construct a deep learning model, which is expected to mine high-level semantic information\cite{huang2021self,huang2021abnormal,huang2022self} of incomplete multi-view features to cope with the data from more complex sources\cite{ngiam2011multimodal,wen2020dimc,wen2020cdimc}. In addition, how to improve the performance and stability of the algorithm on high missing rates and large datasets is also a crucial problem.

	
	%

	\appendices
	



	\ifCLASSOPTIONcaptionsoff
	\newpage
	\fi

	
	
	
	%
	
	\bibliographystyle{IEEEtran}
	\bibliography{ref}
	
	%
	
	
	
	
%
	\begin{IEEEbiography}[{\includegraphics[width=1in,clip,keepaspectratio]{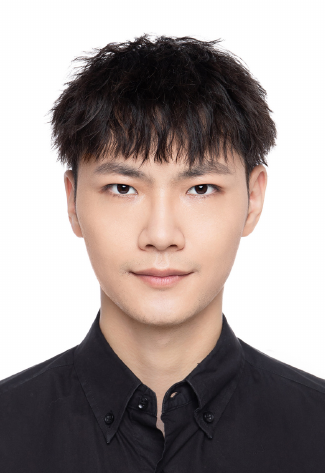}}]{Chengliang Liu}
	(Graduate Student Member, IEEE) received the B.S. degree in computer science from the Jilin University, Changchun, China, in 2018 and the M.S. degree in computer science from the Huazhong University of Science and Technology, Wuhan, China, in 2020. He is currently working toward the Doctoral degree with the Harbin Institute of Technology, Shenzhen. China. His research interests include machine learning and computer vision, especially multiview representation learning.
	\end{IEEEbiography}
	\begin{IEEEbiography}[{\includegraphics[width=1.0in,height=1.2in]{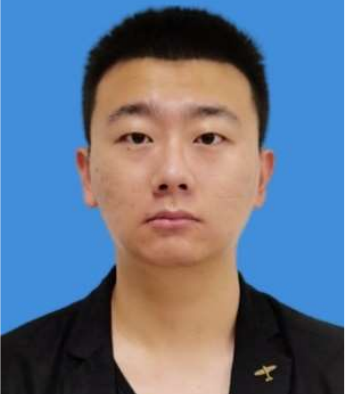}}]{Zhihao~Wu}
	received the B.S. degree in Department of Computer Science from Harbin Engineering University, Harbin, China, in 2019. He is currently pursuing the Ph.D. degree in Harbin Institute of Technology, Shenzhen. His research interests include computer vision and machine learning, especially weakly supervised object detection.
	\end{IEEEbiography}
	\begin{IEEEbiography}[{\includegraphics[width=1.0in,height=1.0in]{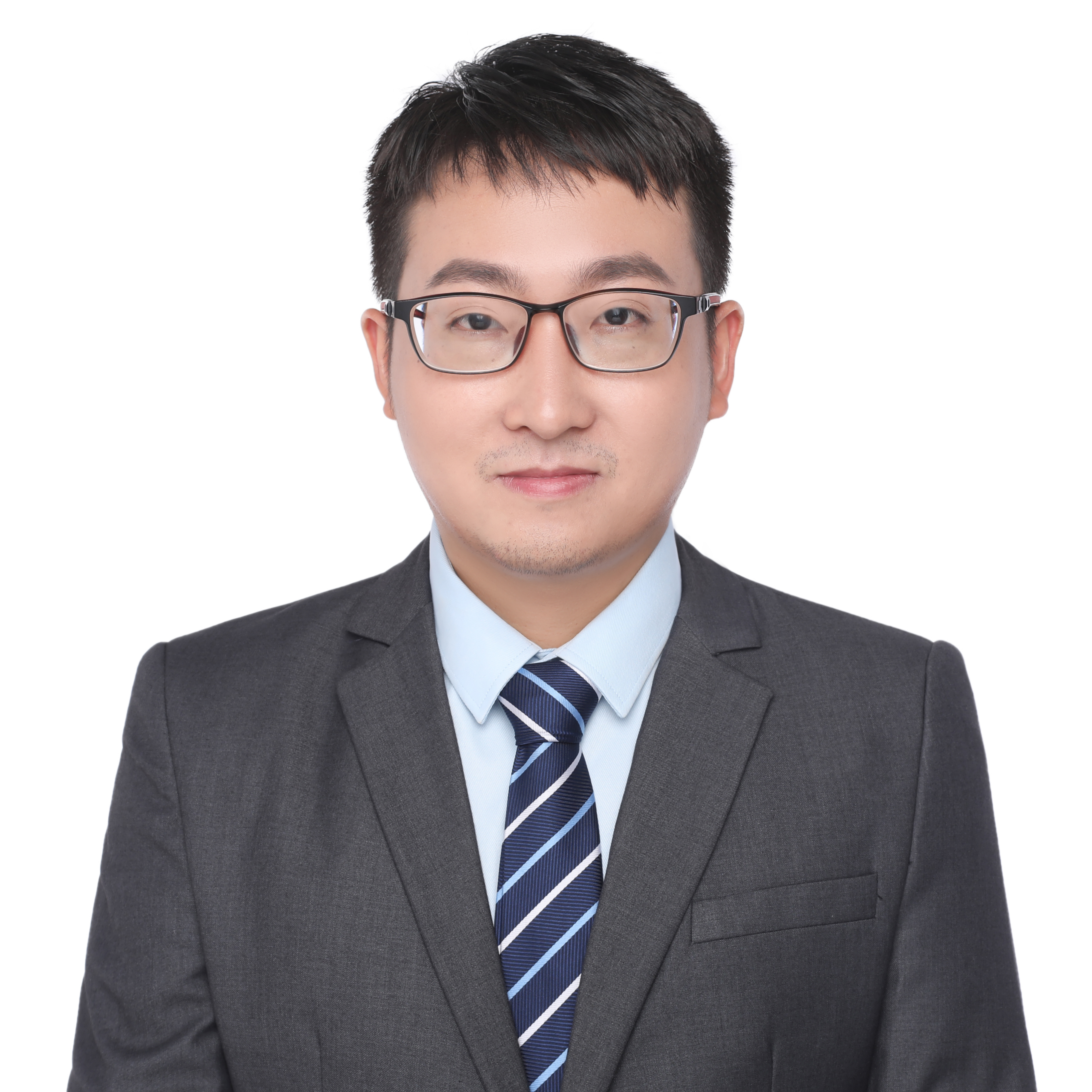}}]{Jie Wen}
	Jie Wen received the Ph.D. degree in Computer Science and Technology at Harbin Institute of Technology, Shenzhen. His research interests include, biometrics, pattern recognition and machine learning. More information please refer to https://sites.google.com/view/jerry-wen-hit/home.
	\end{IEEEbiography}
	
	\begin{IEEEbiography}[{\includegraphics[width=1.0in,height=1.2in]{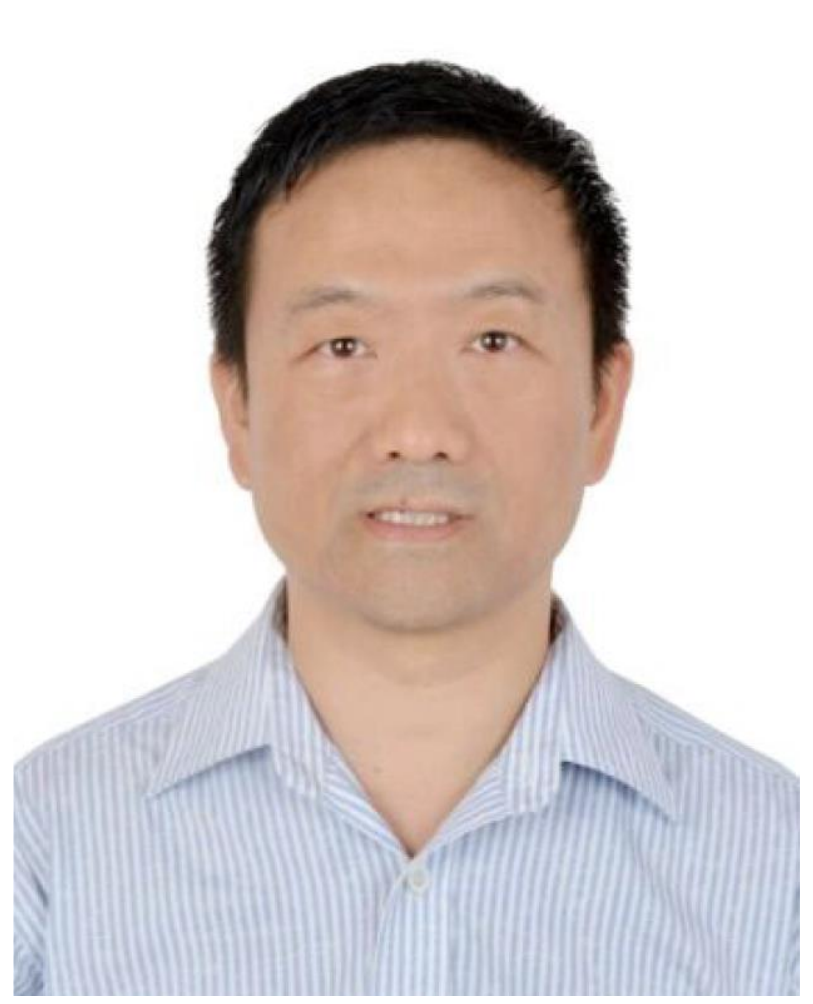}}]{Yong Xu}
	(Senior Member, IEEE) received his B.S. degree, M.S. degree in 1994 and 1997, respectively. He received the Ph.D. degree in Pattern Recognition and Intelligence system at NUST (China) in 2005. He is currently an Professor with the School of Computer Science and Technology, Harbin Institute of Technology (HIT), Shenzhen. His research interests include pattern recognition, deep learning, biometrics, machine learning and video analysis. He has published over 70 papers in toptier academic journals and conferences. His articles have been cited more than 5,800 times in the Web of Science, and 15,000 times in the Google Scholar. He has served as an Co-Editors-in-Chief of the International Journal of Image and Graphics, an Associate Editor of the CAAI Transactions on Intelligence Technology, an editor of the Pattern Recognition and Artificial Intelligence. 
	\end{IEEEbiography}
	\begin{IEEEbiography}[{\includegraphics[width=1.0in,height=1.2in]{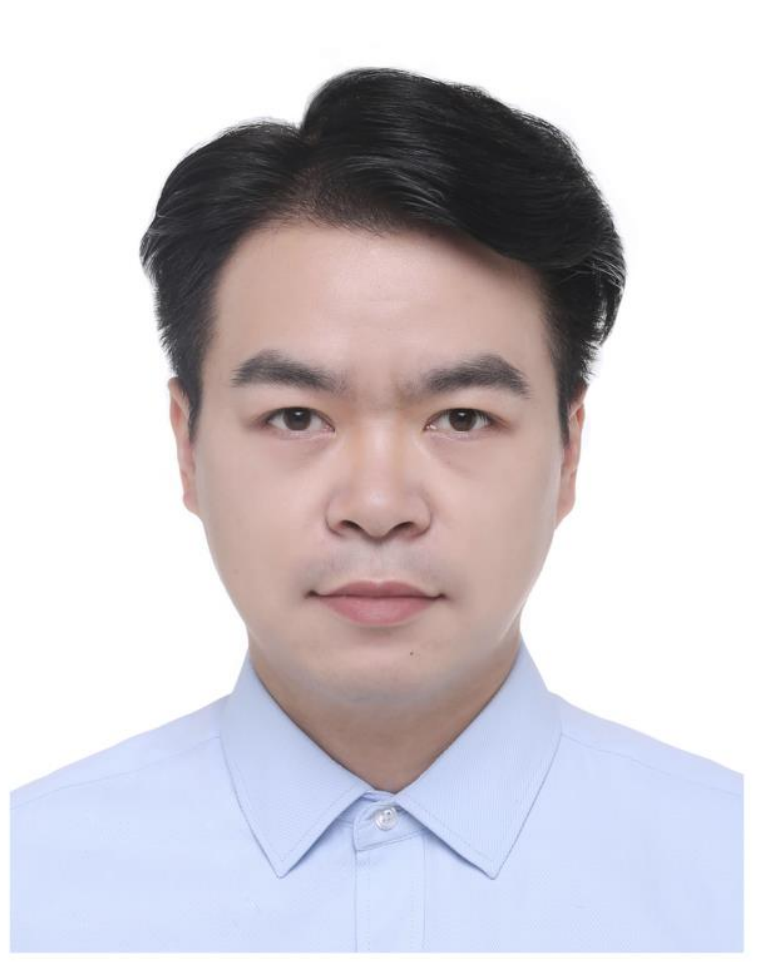}}]{Chao Huang}
		is currently pursuing the Ph.D degree in the School of Computer Science and Technology, Harbin Institute of Technology (HIT), Shenzhen. He received the B.S. degree at Ningbo University, in 2016. He has published over 10 technical papers at prestigious international journals and conferences, including the IEEE TIP, IEEE TNNLS, IEEE TCYB, IEEE TMM, IEEE TII, ACM MM, etc. His research interests include visual anomaly detection, video analysis, object detection, image/video coding and deep learning.
		\end{IEEEbiography}
	\vfill

	
	

\end{document}